\documentclass{article}

\usepackage{arxiv}

\usepackage[utf8]{inputenc} 
\usepackage[T1]{fontenc}    
\usepackage{hyperref}       
\usepackage{url}            
\usepackage{booktabs}       
\usepackage{amsfonts}       
\usepackage{nicefrac}       
\usepackage{microtype}      
\usepackage{lipsum}		
\usepackage[numbers]{natbib}
\usepackage{doi}
\usepackage{color, colortbl}
\usepackage{xcolor} 
\definecolor{Gray}{gray}{0.9}
\definecolor{LightGray}{gray}{0.9}
\usepackage{caption}
\usepackage{subcaption}
\usepackage{graphicx} 
\usepackage{amsmath}
\usepackage{amssymb} 
\usepackage{hyperref} 
\newcommand{\email}[1]{\href{mailto:#1}{\nolinkurl{#1}}}

\title{A tutorial on multi-view autoencoders using the multi-view-AE library}

\author{{\hspace{1mm}Ana ~Lawry Aguila}\\
	University College London\\
    London\\
    WC1E 6BT\\
\email{ana.aguila.18@ucl.ac.uk}\\
	\And
	{\hspace{1mm}Andre ~Altmann} \\
	University College London\\
    London\\
    WC1E 6BT\\
}
\date{}

\renewcommand{\headeright}{Technical Report}
\renewcommand{\shorttitle}{A tutorial on multi-view autoencoders}

\usepackage[frozencache,cachedir=.]{minted}
\begin{document}
\maketitle

\begin{abstract}

There has been a growing interest in recent years in modelling multiple modalities (or views) of data to for example, understand the relationship between modalities or to generate missing data. Multi-view autoencoders have gained significant traction for their adaptability and versatility in modelling multi-modal data, demonstrating an ability to tailor their approach to suit the characteristics of the data at hand. However, most multi-view autoencoders have inconsistent notation and are often implemented using different coding frameworks. To address this, we present a unified mathematical framework for multi-view autoencoders, consolidating their formulations. Moreover, we offer insights into the motivation and theoretical advantages of each model. To facilitate accessibility and practical use, we extend the documentation and functionality of the previously introduced \texttt{multi-view-AE} library. This library offers Python implementations of numerous multi-view autoencoder models, presented within a user-friendly framework. Through benchmarking experiments, we evaluate our implementations against previous ones, demonstrating comparable or superior performance. This work aims to establish a cohesive foundation for multi-modal modelling, serving as a valuable educational resource in the field.

\end{abstract}

\section{Introduction}

Often, data can be naturally described via multiple views or modalities which contain complementary information about the data at hand. These modalities can be modelled jointly using multi-view methods. The joint modelling of multiple modalities has been explored in many research fields such as medical imaging \citep{Serra2018}, chemistry \citep{Sjostrom1983}, and natural language processing \citep{Sadr2020}. 

A popular approach to modelling multiple views are multi-view autoencoders. Generally, this involves learning separate encoder and decoder functions for each modality with the latent representations being combined or associated in some way. The appeal of multi-view autoencoders lies in the versatility of the form of the encoder and decoder functions, ease of extension to handle multiple views, generative properties, and adaptability to large-scale datasets. 

\begin{figure}[ht]
    \centering
    \begin{subfigure}[b]{0.271\textwidth}
         \centering
         \includegraphics[width=\textwidth, trim={0.2cm 0cm 0cm 0cm}, scale=0.5,clip]{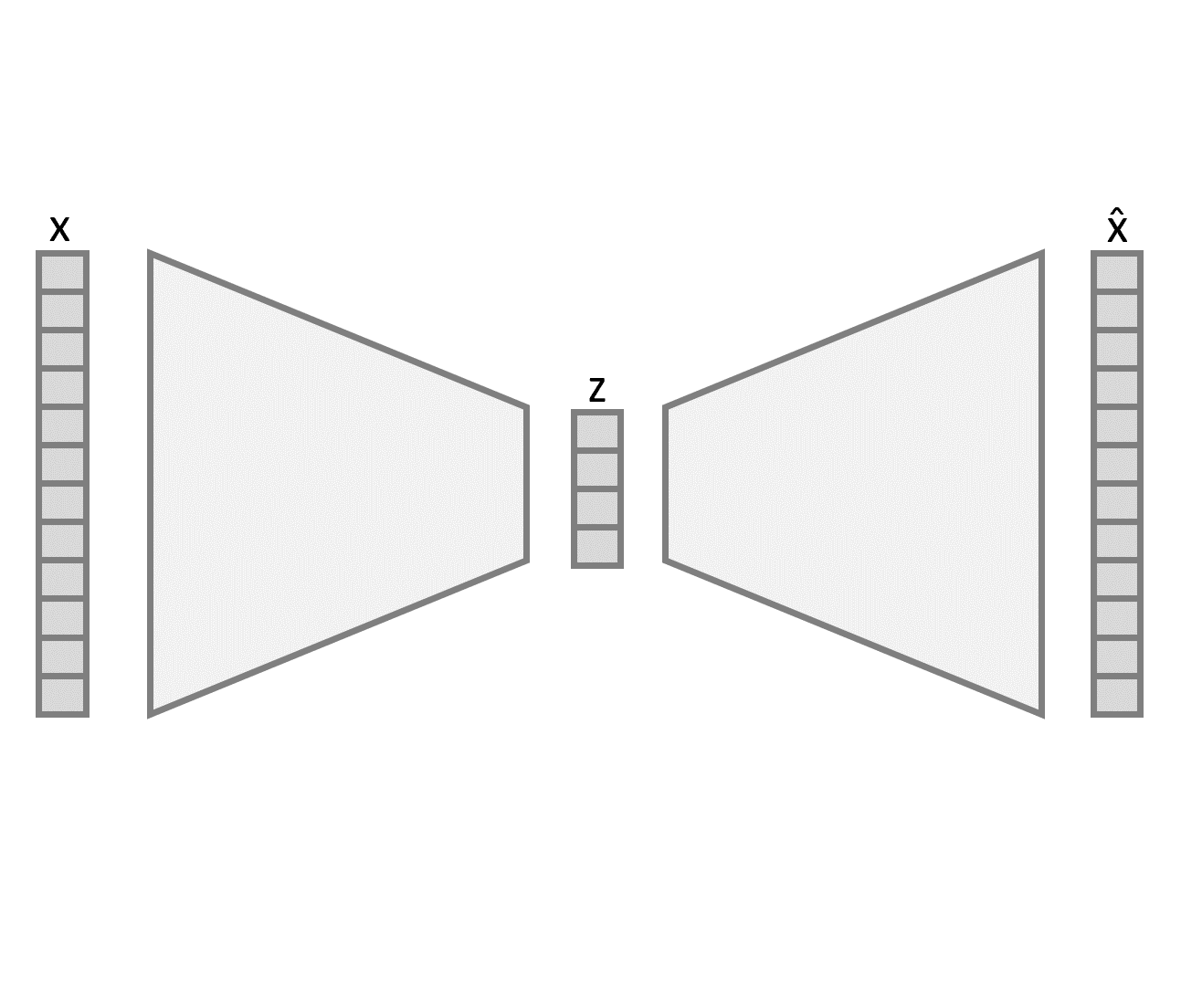}
         \caption{}\label{fig:framework_ae}
     \end{subfigure}
     \hfill
     \begin{subfigure}[b]{0.271\textwidth}
         \centering
         \includegraphics[width=\textwidth, trim={0cm 0cm 0.2cm 0cm}, scale=0.5,clip]{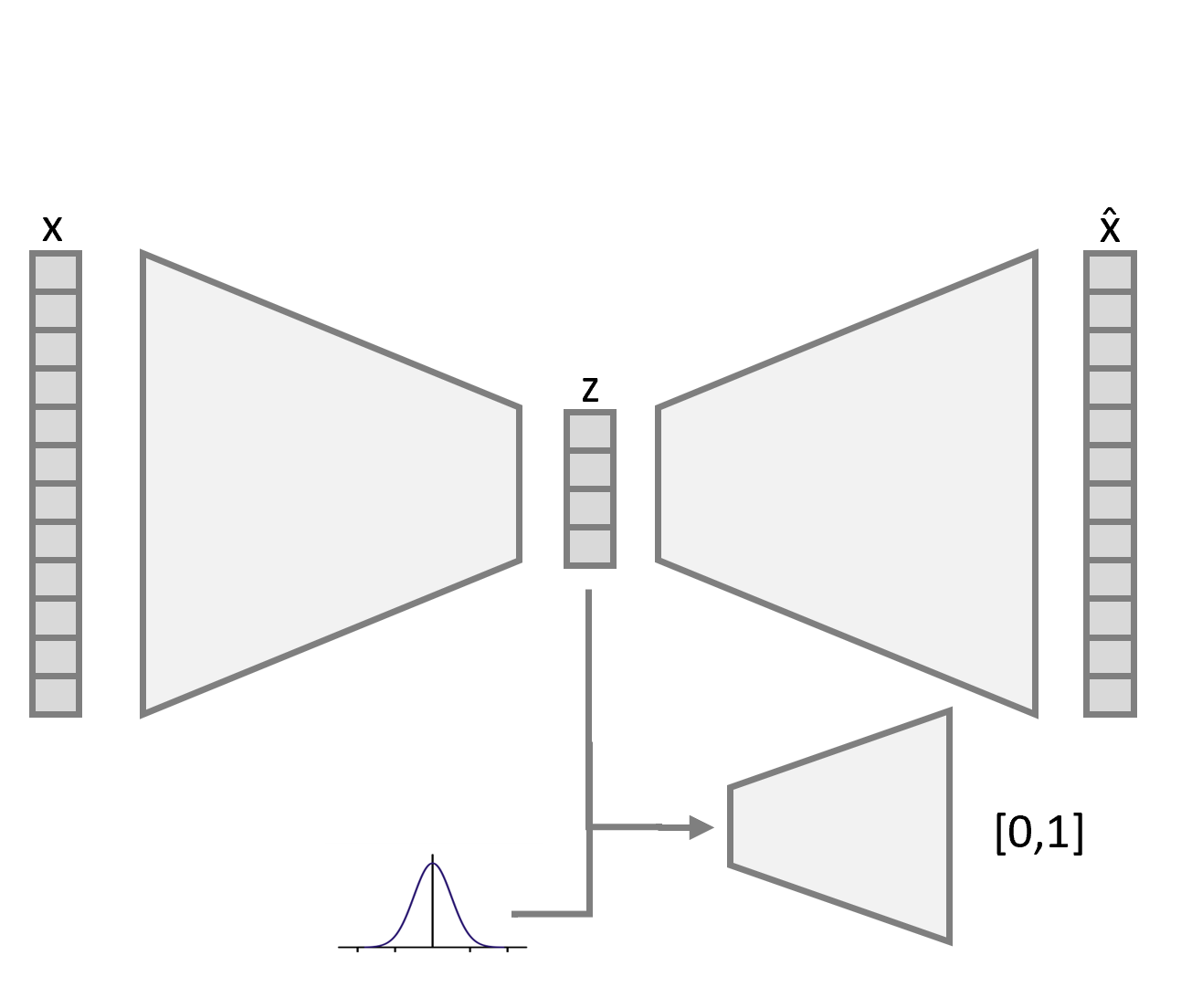}
         \caption{}\label{fig:framework_aae}
     \end{subfigure}
     \hfill
     \begin{subfigure}[b]{0.358\textwidth}
         \centering
         \includegraphics[width=\textwidth, trim={0.2cm 0.32cm 0cm 0cm}, scale=0.5,clip]{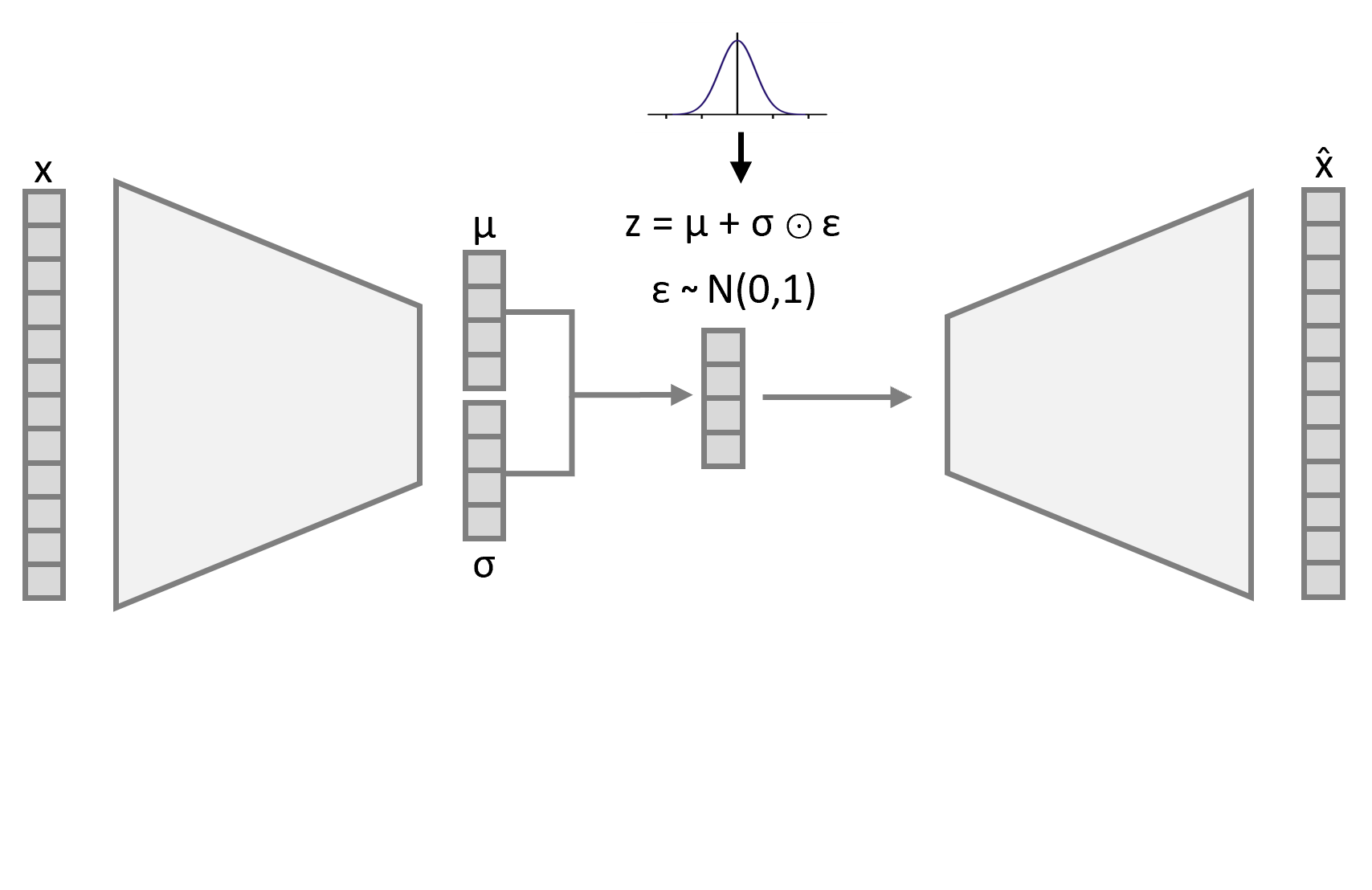}
         \caption{}\label{fig:framework_vae}
     \end{subfigure}
        \caption{Single view autoencoder frameworks; (a) vanilla autoencoder, (b) Adversarial Autoencoder, (c) Variational Autoencoder.}\label{fig:Autoencodermodels}
\end{figure}

Autoencoders often require some form of regularisation to learn more useful representations and reduce the generalisation error. In Variational Autoencoders (VAEs), the latent space is regularised by mapping the encoding distribution to a gaussian prior using a Kullback–Leibler (KL) divergence term. Among multi-view autoencoder models, the most prevalent are the multi-view extensions of VAEs. However, alternative frameworks exist, such as multi-view Adversarial Autoencoders (AAEs) \citep{Wang2019b}. In AAEs, the latent space is regularised by mapping the encoding distribution to a prior using an auxiliary discriminator tasked with distinguishing samples from the posterior and prior distributions. The choice of multi-view autoencoder model may be influenced by various factors of the application process, such as stability during training. Figure \ref{fig:Autoencodermodels} shows the frameworks for uni-modal vanilla, adversarial, and variational autoencoders.

\begin{figure}[ht]
     \centering
    \begin{subfigure}[b]{0.246\textwidth}
         \centering
         \includegraphics[width=\textwidth, trim={0cm 0cm 0cm 0cm}, scale=0.5]{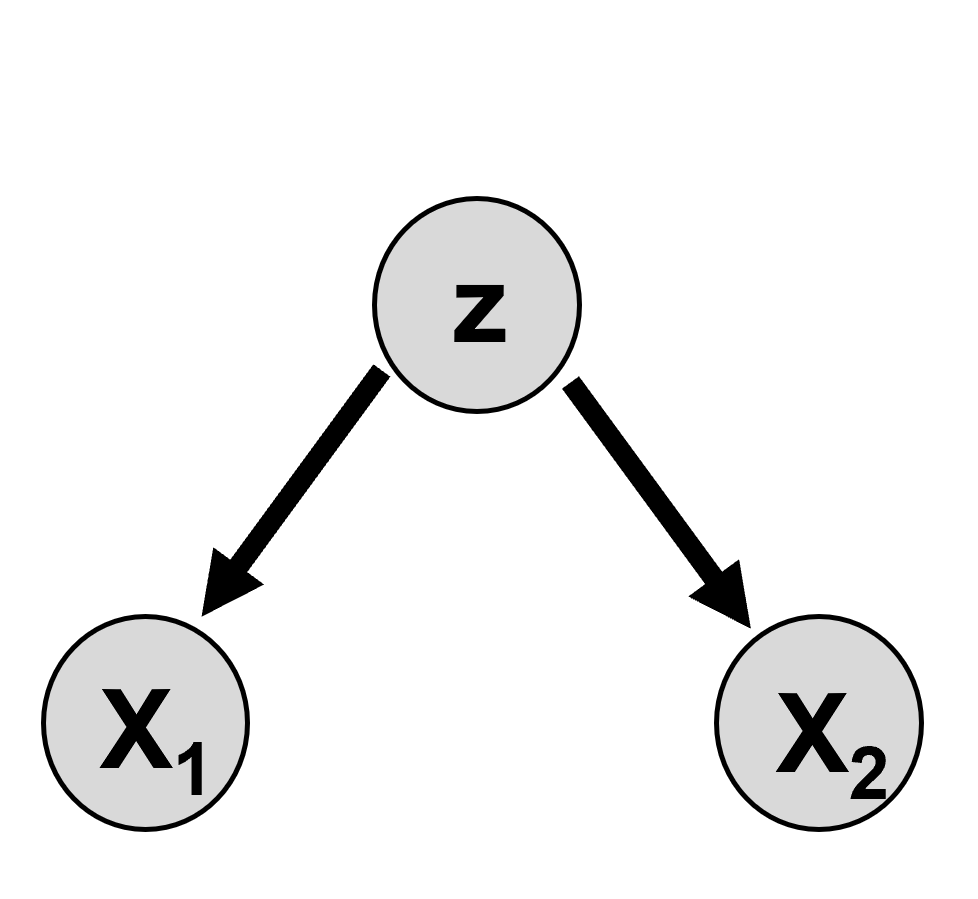}
         \caption{}\label{fig:AElatent_a}
     \end{subfigure}
     \hfill
     \begin{subfigure}[b]{0.246\textwidth}
         \centering
         \includegraphics[width=\textwidth, trim={0cm 0cm 0cm 0cm}, scale=0.5]{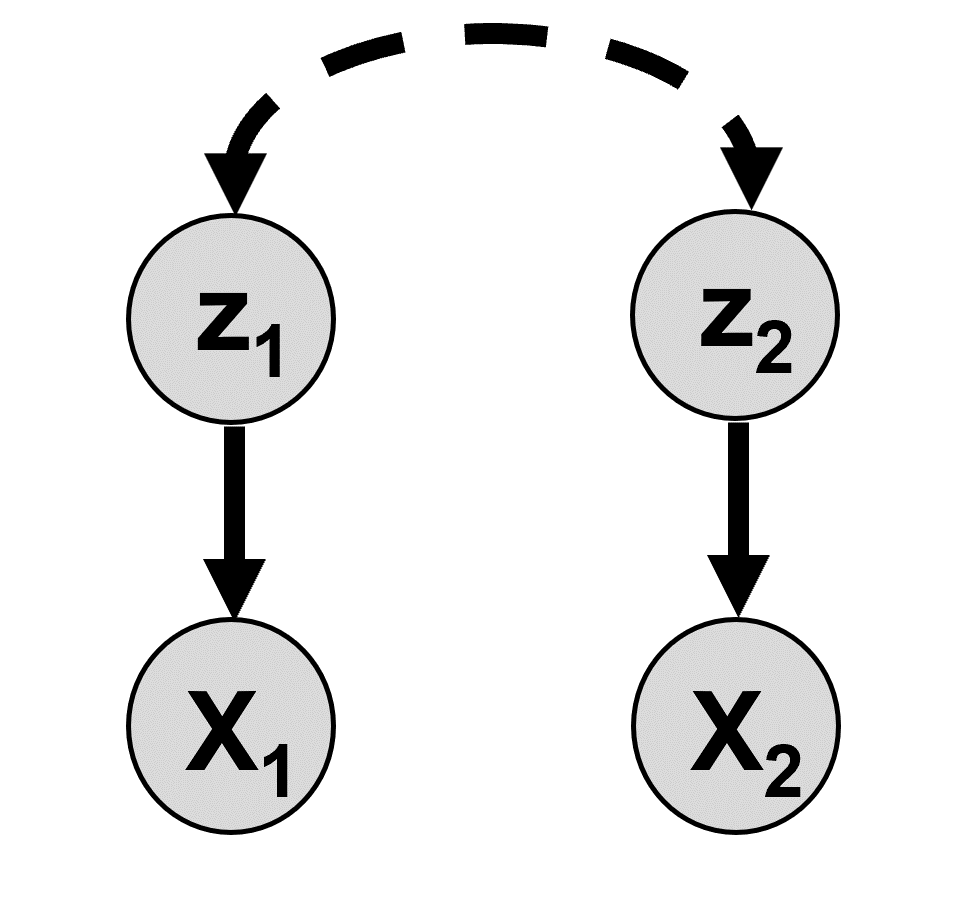}
         \caption{}\label{fig:AElatent_b}
     \end{subfigure}
     \hfill
     \begin{subfigure}[b]{0.408\textwidth}
         \centering
         \includegraphics[width=\textwidth, trim={0cm 0.32cm 0cm 0cm}, scale=0.5]{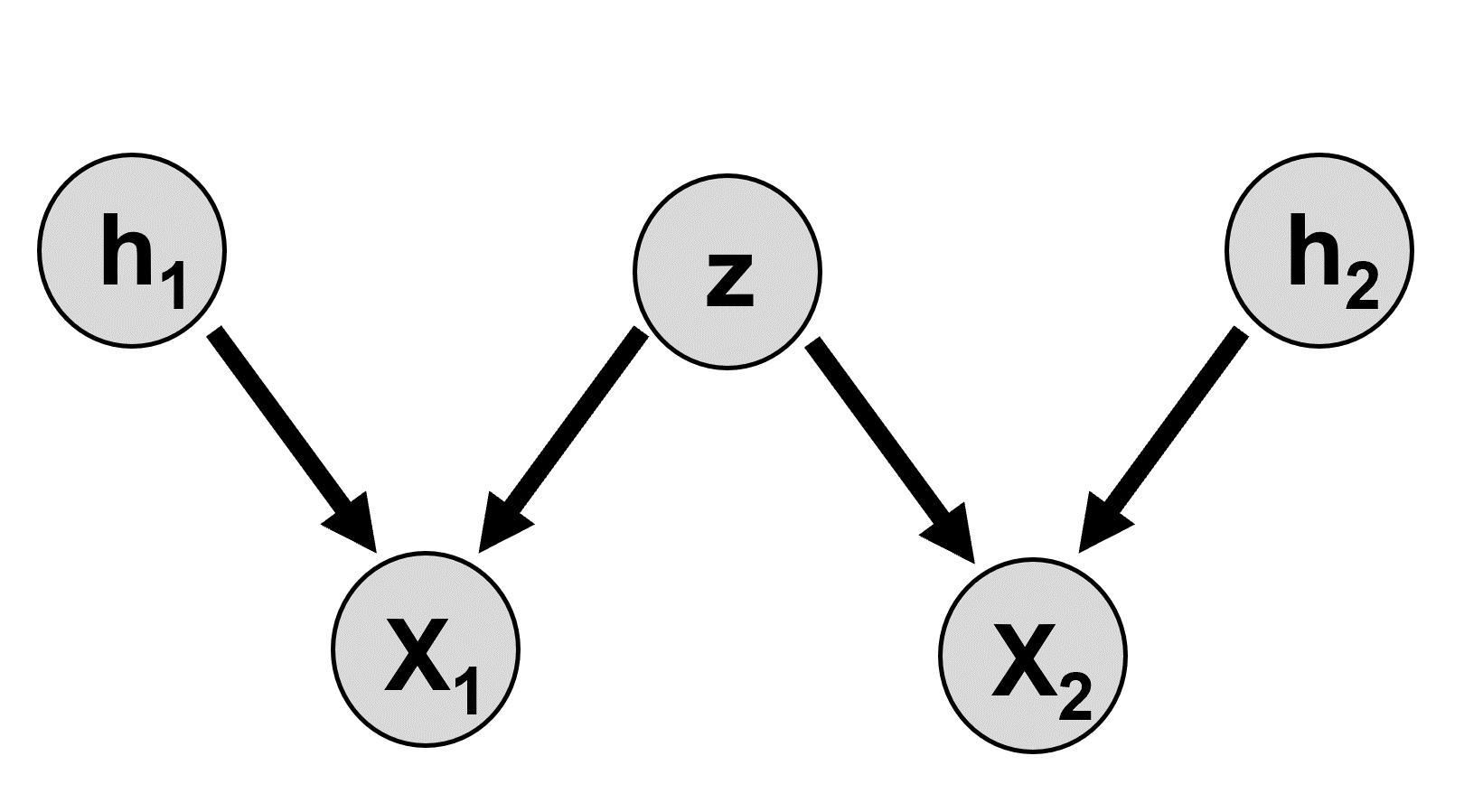}
         \caption{}\label{fig:AElatent_c}
     \end{subfigure}
    \caption{Latent variable models for two input views. Latent variable model where data $\textbf{X}_1$ and $\textbf{X}_2$ (a) share an underlying latent factor $\textbf{z}$, (b) have associated latent factors $\textbf{z}_1$ and $\textbf{z}_2$ and (c) share an underlying latent factor as well as view specific private latent variables.}\label{fig:AElatent}
\end{figure}

\begin{figure}[ht]
     \centering
     \begin{subfigure}[b]{0.32\textwidth}
         \centering
         \includegraphics[width=\textwidth, trim={0.2cm 0cm 0cm 0cm}, scale=0.5, clip]{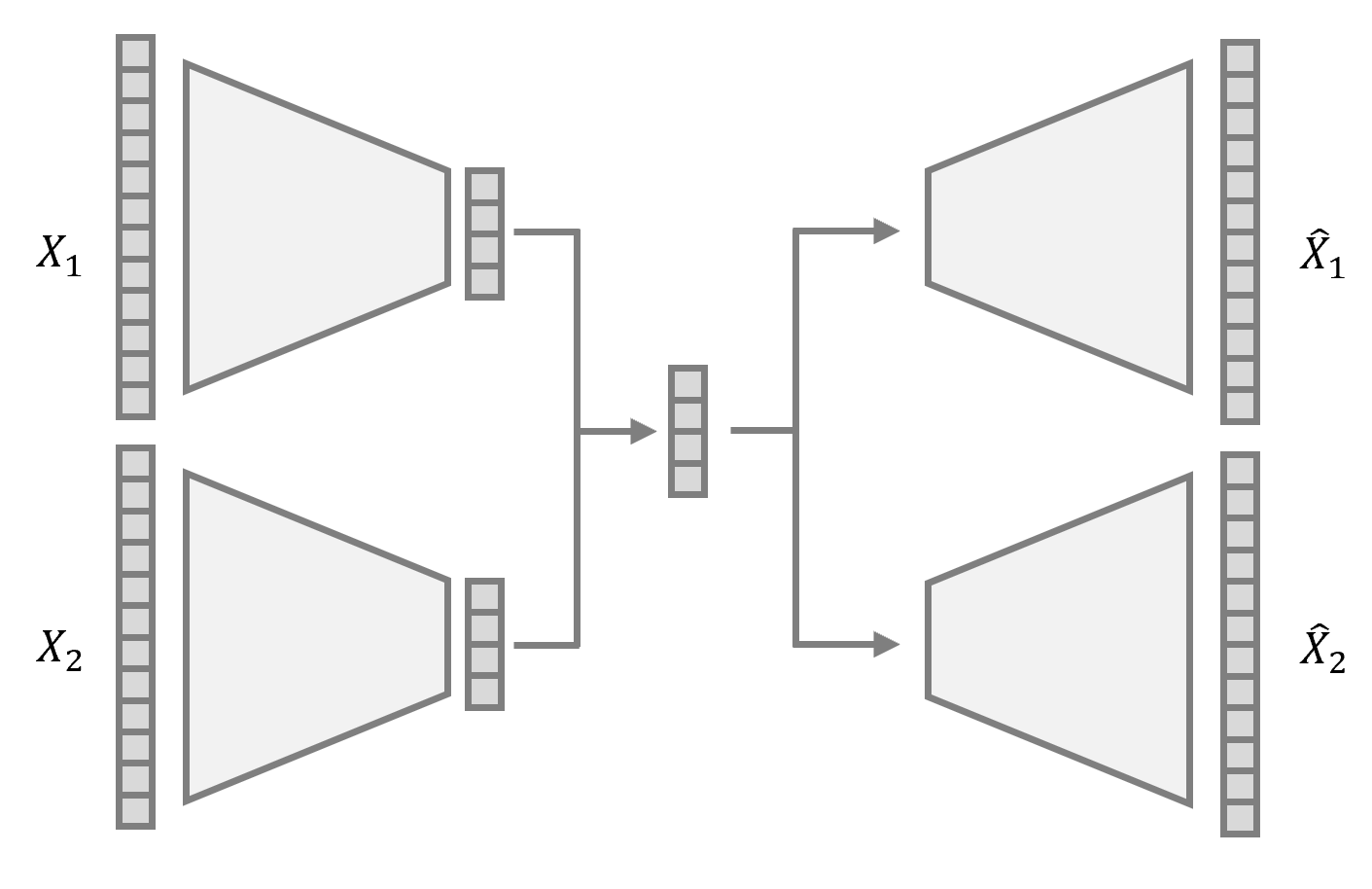}
         \caption{}
     \end{subfigure}
     \hfill
     \begin{subfigure}[b]{0.32\textwidth}
         \centering
         \includegraphics[width=\textwidth, trim={0.2cm .1cm 0cm 0cm}, scale=0.5, clip]{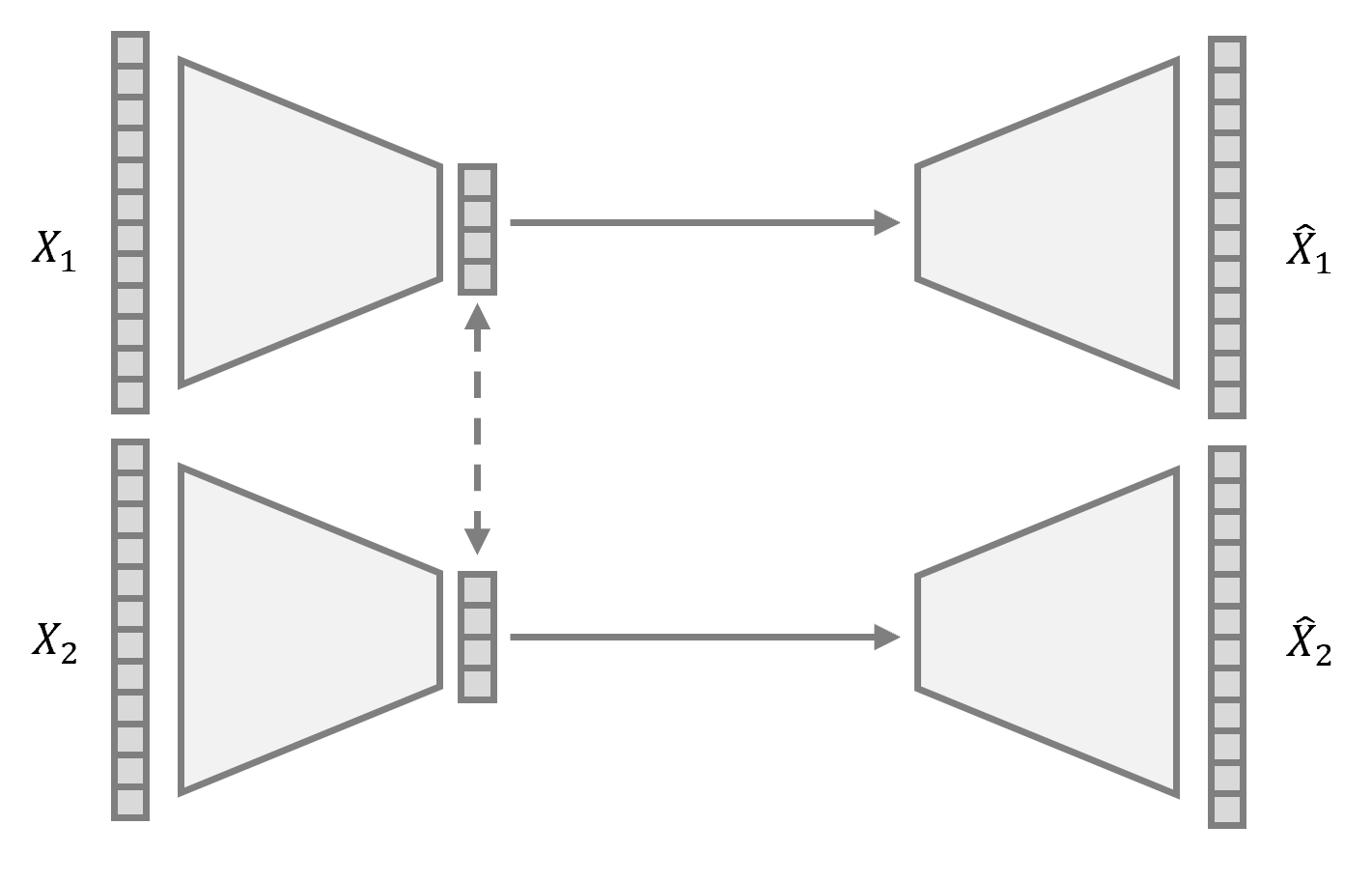}
         \caption{}
     \end{subfigure}
     \hfill
     \begin{subfigure}[b]{0.32\textwidth}
         \centering
         \includegraphics[width=\textwidth, trim={0.2cm 0.1cm 0cm 0cm}, scale=0.5,clip]{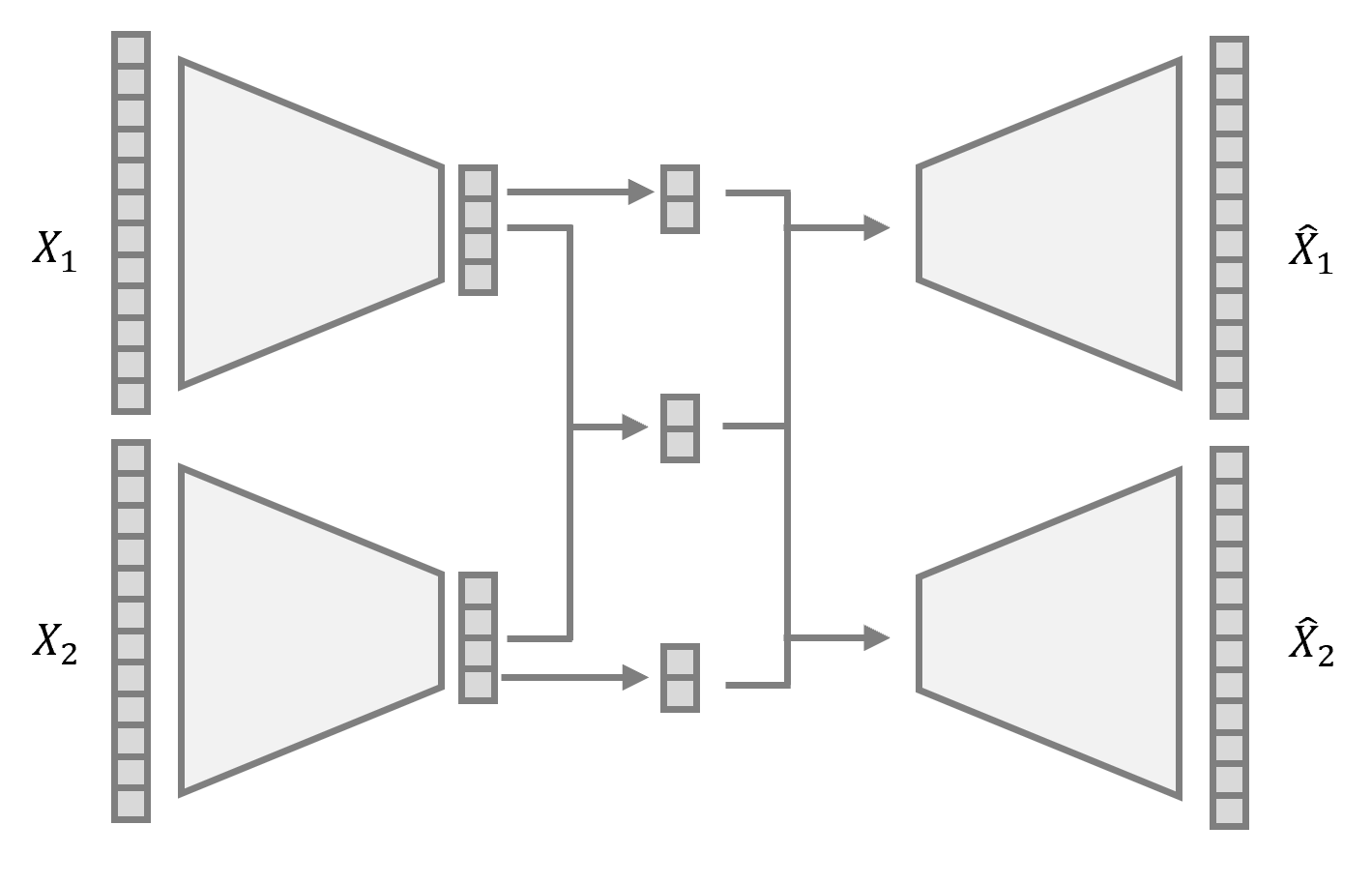}
         \caption{}
     \end{subfigure}
    \caption{Example frameworks of a two-view autoencoder for data $\mathbf{X}_1$ and $\textbf{X}_2$ for (a) a joint model, where the individual latent spaces are combined and the reconstruction is carried out from the joint latent space, (b) a coordinated model, where the latent representations are coordinated by an addition loss term for association between the latent variable, and (c) a joint model with shared and private latent variables.}\label{fig:AEexample}
\end{figure}

Even within these regularisation frameworks there are vast modelling differences to be considered when choosing the best model for the task at hand. Figure \ref{fig:AElatent} depicts three possible latent variable models for modelling two views of data; $\mathbf{X}_1$ and $\textbf{X}_2$. Figure \ref{fig:AElatent_a} shows the joint latent variable model \citep{Suzuki2022} where both views, $\mathbf{X}_1$ and $\textbf{X}_2$, share an underlying factor. The latent variable model in Figure \ref{fig:AElatent_b} shows a coordinated model \citep{Suzuki2022}, which assumes some relationship between the latent variables, $\mathbf{z}_{1}$ and $\mathbf{z}_{2}$ of $\mathbf{X}_1$ and $\textbf{X}_2$ respectively. Figure \ref{fig:AElatent_c} shows a latent variable model where views share a latent variable $\mathbf{z}$ as well as view specific private latent variables $\mathbf{h}_{1}$ and $\mathbf{h}_{2}$. Example multi-view autoencoder frameworks built for these three latent variable models are given in Figure \ref{fig:AEexample}.

Which latent variable model is most appropriate depends on the desired outcome of the learning task. Even for a particular latent variable model, several modelling considerations can significantly influence the structure of the multi-view autoencoder framework. One such consideration is the choice of function to use to learn the joint representation $\mathbf{z}$. This modelling consideration alone has given rise to a number of different multi-modal VAEs \citep{Wu2018,Shi2019,Suzuki2016,Sutter2021,Sutter2021b,Lawryaguila2023}. In the following sections, we introduce several multi-view autoencoder models through the lens of the \texttt{multi-view-AE} package.

\section{Package benefits}

There exist many different multi-view autoencoder frameworks with the best method of choice depending on the specific task. Existing code is often implemented using different Deep Learning frameworks or varied programming styles making it difficult for users to compare methods. Given the large number of multi-view autoencoders and versatility of architecture, it is important to consider which model would best suit the use case. \texttt{multi-view-AE} is a Python library which implements several variants of multi-view autoencoders in a simple, flexible, and easy-to-use framework \citep{LawryAguila_software2023}. 

The motivation for developing the \texttt{multi-view-AE} package was to widen the accessibility of these algorithms by allowing users to easily test methods on new datasets and enable developers to compare methods and extend code for new research applications. We would like to highlight the following benefits of the package.

\begin{itemize}
    \item \textbf{Simple API and code structure.} The \texttt{multi-view-AE} package is implemented with a similar interface to \texttt{scikit-learn} \citep{sklearn_api2013} with common and straight-forward functions implemented for all models. This makes it simple for users to train and evaluate models without requiring detailed knowledge of the methodology.
    \item \textbf{Modular structure.} The modular structure of the \texttt{multi-view-AE} package allows developers to choose which element of the code to extend and swap out existing Python classes for new implementations whilst leaving the wider codebase untouched.
    \item \textbf{Suitable for beginners and developers.} For most of the building blocks of the multi-view autoencoders, the users have the flexibility to choose the class (e.g. the encoder or decoder network) from the available implementations, or to contribute their own. As such, the \texttt{multi-view-AE} package is accessible to both beginners, with off-the-shelf models, and experts, who wish to adapt the existing framework for further research purposes.
    \item \textbf{Use of state-of-the-art deep learning toolkits.} The \texttt{multi-view-AE} package uses the \texttt{PyTorch-Lightning} API which offers the same functionality as raw \texttt{PyTorch} in a more structured and streamlined way. This offers users more flexibility, faster training and optimisation time, and high scalability.
    
\end{itemize}

There are a few libraries related to \texttt{multi-view-AE}. The \texttt{Pixyz} library \citep{Suzuki2021}, for example, implements a number of multi-view autoencoder methods. However, \texttt{Pixyz} is designed for the wider field of deep generative modelling whereas \texttt{multi-view-AE} focuses specifically on multi-view autoencoder models. As such \texttt{multi-view-AE} builds upon \texttt{Pixyz}'s multi-view offering providing a wider range of multi-view methods. Since the initial publication of \texttt{multi-view-AE} \cite{LawryAguila_software2023}, the \texttt{MultiVae} library was introduced which implements many of the multi-modal VAEs implemented in the \texttt{multi-view-AE} library \citep{senellart2023}. However, this library does not implement as wide a selection of models as \texttt{multi-view-AE} and, importantly, neither the paper nor the accompanying code for \texttt{MultiVae} has undergone peer review.

\section{Implemented models}

A complete list of models implemented, at the time of writing, in the \texttt{multi-view-AE} package is given in Table \ref{tab:models}. We note that the models implemented in the \texttt{multi-view-AE} package will differ slightly from those implemented in the original paper due to differences in deep learning libraries, function components, and other design choices. However, the models make the same mathematical and framework choices as the original implementations. For a number of models, we have benchmarked our models against previously published results using toy datasets. 

\begin{table}[!ht]
\resizebox{\textwidth}{!}{%
\begin{tabular}{|l|p{13cm}|l|}
    \hline
    \textbf{Model class} & \textbf{Model name} & \textbf{No. views}\\
    \hline
    AE & Multi-view Autoencoder & >=1\\
    \hline
    JMVAE  & Joint Multi-modal Variational Autoencoder (JMVAE-kl) \citep{Suzuki2016} & 2 \\
    \hline
    DCCAE  & Deep Canonically Correlated Autoencoder \citep{Wang2016} & 2 \\
    \hline
    DVCCA  & Deep Variational CCA \citep{Wang2016a} & 2\\
    \hline
    mcVAE & Multi-Channel Variational Autoencoder (mcVAE) \citep{Antelmi2019} & >=1 \\
    \hline
    mVAE  & Multi-modal Variational Autoencoder (MVAE) \citep{Wu2018} & >=1 \\
    \hline
    me\_mVAE  & Multi-modal Variational Autoencoder (MVAE) with separate ELBO terms for each view \citep{Wu2018} & >=1\\
    \hline
    mmVAE   & Variational Mixture-of-Experts Autoencoder (MMVAE) \citep{Shi2019} & >=1 \\
    \hline
    MVTCAE  & Multi-View Total Correlation Autoencoder (MVTCAE) \citep{Hwang2021} & >=1 \\
    \hline
    MoPoEVAE  & Mixture-of-Products-of-Experts VAE \citep{Sutter2021} & >=1\\
    \hline
    mmJSD  & Multi-modal Jensen-Shannon divergence model (mmJSD) \citep{Sutter2021b} & >=1 \\
    \hline
    weighted\_mVAE & Generalised Product-of-Experts Variational Autoencoder (gPoE-MVAE) \citep{Lawryaguila2023} & >=1\\
    \hline
    mmVAEPlus & Mixture-of-Experts multi-modal VAE Plus (MMVAE+) \citep{Palumbo2022} & >=1\\
    \hline
    DMVAE & Disentangled multi-modal Variational Autoencoder \citep{Lee2021} & >=1 \\
    \hline
    mAAE  & Multi-view Adversarial Autoencoder & >=1\\
    \hline
    mWAE  & Multi-view Adversarial Autoencoder with Wasserstein loss & >=1\\
    \hline
\end{tabular}}
\caption{Models implemented in the \texttt{multi-view-AE} package.}
\label{tab:models} 
\end{table}

In the following sections we will introduce the objective functions and key modelling features of each implemented model. We let $\textbf{X}=\{\textbf{x}_m\}^M_{m=1}$ be an observation of $M$ modalities for a single sample. Where a model is restricted to two input modalities, we use $\mathbf{x}_1 \in \textbf{X}_1$ and $\mathbf{x}_2 \in \textbf{X}_2$ to indicate that the data vectors of an individual subject come from the data matrices $\textbf{X}_1$ and $\textbf{X}_2$. We use the same notation across models to indicate common distributions and parameters. We would like to note that this in itself is a contribution, as previous work has often used different notation making it difficult to easily compare model formulation. 

Most of the models implemented in the \texttt{multi-view-AE} package are generative models (and mostly multi-modal VAEs) which assume the latent variable models given in Figures \ref{fig:AElatent_a} and \ref{fig:AElatent_c} (for $M=2$). We can write Figure \ref{fig:AElatent_a} in equation form:

\begin{equation}
p(\textbf{z} \mid \textbf{X})=\frac{p(\textbf{z}) p(\textbf{X} \mid \textbf{z})}{p(\textbf{X})}
\end{equation}
with a multi-modal generative model given by;
\begin{equation*}
p_\theta(\textbf{X}, \textbf{z})=p(\textbf{z}) \prod_{m=1}^M p_{\theta_m}\left(\textbf{x}_m \mid \textbf{z}\right)
\end{equation*} 
where we assume modalities are conditionally independent given a common latent variable. Similarly, we can write Figure \ref{fig:AElatent_c},  where modalities share a common latent variable, $\textbf{z}$, as well as conditionally independent modality specific latent variables, $\textbf{h}$, in equation form: 

\begin{equation}\label{eq:Bayes_theorem_multi-view_private}
p(\textbf{z}, \textbf{h} \mid \textbf{X})=\frac{p(\textbf{z})p(\textbf{h}) p(\textbf{X} \mid \textbf{z}, \textbf{h})}{p(\textbf{X})}.
\end{equation} 

Notice that following the latent variable model given in Figure \ref{fig:AElatent_c}, $ p(\textbf{X} \mid \textbf{z}, \textbf{h})=\prod_{m=1}^M p_{\theta_m}\left(\textbf{x}_m \mid \textbf{z}, \textbf{h}_m\right)$ and hence from Equation \ref{eq:Bayes_theorem_multi-view_private} we obtain a multi-modal generative model given by;

\begin{equation}
p_\theta(\textbf{X}, \textbf{z}, \textbf{h})=p(\textbf{z}) \prod_{m=1}^M p\left(\textbf{h}_m\right) \prod_{m=1}^M p_{\theta_m}\left(\textbf{x}_m \mid \textbf{z}, \textbf{h}_m\right).
\end{equation} 

We will use the formulas given here as the foundations for describing the objective function of each model implemented in the \texttt{multi-view-AE} package.

\subsection*{Vanilla autoencoder (AE)}

A classical autoencoder can be extended to multiple modalities by training an encoder, $f_e$, and decoder, $f_d$, network per modality. The multi-modal autoencoder objective function is given by:
\begin{equation}\label{eq:mAE}
\mathcal{L}^{\text{AE}} = \frac{1}{M} \sum_{m=1}^M \frac{1}{M} \sum_{n=1}^M \|\textbf{x}_m-f_d^m(f_e^n(\textbf{x}_n))\|_2^2
\end{equation}
which consists of same-modality ($n=m$) and cross-modality ($n!=m$) reconstruction losses. As such, the objective encourages good reconstruction of each modality from all the modality-specific latent representations. As with the classical autoencoder, the lack of regularisation means the multi-modal autoencoder will likely generalise poorly to unseen data. 

\subsection*{Joint Multi-modal Variational Autoencoder (JMVAE)}

The JMVAE model, proposed by \citet{Suzuki2016}, was one of the first multi-modal VAE. JMVAE assumes a latent variable model illustrated in Figure \ref{fig:AElatent_a} and considers a dataset of two modalities $\textbf{X}_1$ and $\textbf{X}_2$ with joint posterior distribution $q_{\mathbf{\phi}}(\mathbf{z} \mid \textbf{x}_1, \textbf{x}_2)$ where $\phi$ is parameterised by a joint encoder network. The JMVAE objective is given by the following ELBO:
\begin{equation}\label{eq:JMVAE}
\begin{aligned}
&\mathcal{L}^{\text{JMVAE}} = -D_{K L}\left(q_{\phi}(\mathbf{z} \mid \mathbf{x}_1, \mathbf{x}_2) \| p(\mathbf{z})\right)+\\
&\mathbb{E}_{q_{\phi}(\mathbf{z} \mid \mathbf{x}_1)}\left[\log p_{\mathbf{\theta}_1}(\mathbf{x}_1 \mid \mathbf{z})\right]+\mathbb{E}_{q_{\phi}(\mathbf{z} \mid \mathbf{x}_2)}\left[\log p_{\mathbf{\theta}_2}(\mathbf{x}_2 \mid \mathbf{z})\right]
\end{aligned}
\end{equation}
where $p_{\mathbf{\theta}_1}(\mathbf{x}_1 \mid \mathbf{z})$ and $p_{\mathbf{\theta}_2}(\mathbf{x}_2 \mid \mathbf{z})$ are the decoder networks for modalities $\textbf{X}_1$ and  $\textbf{X}_2$ respectively. The first term in the objective function is the KL divergence between the joint posterior distribution, $q_{\mathbf{\phi}}(\mathbf{z} \mid \textbf{x}_1, \textbf{x}_2)$, and the prior, $p(\mathbf{z})$. The second two terms are the expected log-likelihood under the approximate posterior distribution of each modality $\textbf{x}_m$, which corresponds to the negative of the reconstruction loss obtained by decoding from the joint latent representation $\textbf{z}$ using the decoders $p_{\mathbf{\theta}_1}(\mathbf{x}_1 \mid \mathbf{z})$ and $p_{\mathbf{\theta}_2}(\mathbf{x}_2 \mid \mathbf{z})$. 

\citet{Suzuki2016} also introduce JMVAE-kl which includes an additional two single input encoding networks, $q_{\mathbf{\phi}_{\textbf{x}_1}}(\mathbf{z} \mid \textbf{x}_1)$ and $q_{\mathbf{\phi}_{\textbf{x}_2}}(\mathbf{z} \mid \textbf{x}_2)$. The JMVAE-kl objective function includes additional KL divergence terms to encourage the joint encoding distribution to be similar to the separate encoders:
\begin{equation}\label{eq:JMVAE_kl}
\begin{aligned}
&\mathcal{L}^{\text{JMVAE-kl}} = \mathcal{L}^{\text{JMVAE}} - \\
&\alpha \left[ D_{K L}\left(q_{\phi}(\mathbf{z} \mid \mathbf{x}_1, \mathbf{x}_2) \| q_{\mathbf{\phi}_{x_1}}(\mathbf{z} \mid \textbf{x}_1)\right) + 
D_{K L}\left(q_{\phi}(\mathbf{z} \mid \mathbf{x}_1, \mathbf{x}_2) \| q_{\mathbf{\phi}_{x_2}}(\mathbf{z} \mid \textbf{x}_2)\right)
\right] 
\end{aligned}
\end{equation}
where $\alpha$ is a factor that regulates the KL divergence terms. The additional KL terms can be thought of as minimising the variation of information (VI) between the two single encoders \citep{Suzuki2016} and ensuring that a single modality does not dominate the latent space. In the \texttt{multi-view-AE} package, we provide an implementation of the JMVAE-kl model. However, the user can opt for the JMVAE model by setting the $\alpha$ parameter to zero. 

\subsection*{Deep Canonically Correlated Autoencoders (DCCAE)}

Canonical Correlation Analysis (CCA) is a popular linear coordinated model which maximises the correlation between latent variables. The latent variables are given by the projection of the original data matrices onto a pair of singular vectors known as weights. \citet{Andrew2013} developed Deep Canonical Correlation Analysis (DCCA) a deep learning alternative to CCA to learn non-linear mappings of multi-modal information. Deep CCA computes representations by passing two views through functions $f_1$ and $f_2$ with parameters $\mathbf{\theta}_1$ and $\mathbf{\theta}_2$, respectively, which can be learnt by multi-layer neural networks. The parameters are optimised by maximising the correlation between the learnt representations $f_1(\mathbf{X}_1;\mathbf{\theta}_1)$ and $f_2(\mathbf{X}_2;\mathbf{\theta}_2)$:

\begin{equation}\label{eq:DCCA}
    \mathcal{L}^{\text{DCCA}} = -\textrm{ Corr}\left(f_{1}\left(\mathbf{X}_{1} ; \mathbf{\theta}_{1}\right), f_{2}\left(\mathbf{X}_{2} ; \mathbf{\theta}_{2}\right)\right).
\end{equation}
Deep Canonically Correlated Autoencoders (DCCAE) \citep{Wang2016} extend DCCA to an autoencoder combining the DCCA objective with the autoencoder framework:
\begin{equation}\label{eq:DCCAE}
    \mathcal{L}^{\text{DCCAE}} = \mathcal{L}^{\text{DCCA}} + \frac{\lambda}{N} \sum_{i=1}^N  \left( \|\textbf{x}^i_1 - d_1(f_1(\textbf{x}^i_1))\|_2^2 + \|\textbf{x}^i_2 - d_2(f_2(\textbf{x}^i_2))\|_2^2\right)
\end{equation}
where $d_1$ and $d_2$ are the decoder networks for modalities $\textbf{X}_1$ and $\textbf{X}_2$, $N$ is the number of samples, and $\lambda$ is a weighting parameter between the autoencoder and CCA terms. Note that, the DCCAE objective produces sub-optimal results for mini-batch optimisation and is therefore best suited for batch optimisation \citep{Andrew2013}.

\subsection*{Deep Variational CCA (DVCCA)}

\citet{Wang2016a} introduced deep variational CCA (DVCCA) which extends the probabilistic CCA framework \citep{Bach2005} to a non-linear generative model. In a similar approach to VAEs, deep VCCA uses variational inference to approximate the posterior distribution and derives the following ELBO:
\begin{equation}\label{eq:VCCA_loss}
   \mathcal{L}^{\text{VCCA}} = -D_{K L}\left(q_{\phi}(\mathbf{z} \mid \mathbf{x}_1) \| p(\mathbf{z})\right)+\mathbb{E}_{q_{\phi}(\mathbf{z} \mid \mathbf{x}_1)}\left[\log p_{\mathbf{\theta}_1}(\mathbf{x}_1 \mid \mathbf{z})+\log p_{\mathbf{\theta}_2}(\mathbf{x}_2 \mid \mathbf{z})\right]
\end{equation}
where the approximate posterior, $q_{\phi}(\mathbf{z} \mid \mathbf{x}_1)$, and likelihood distributions, $p_{\mathbf{\theta}_1}(\mathbf{x}_{1} \mid \mathbf{z})$ and $p_{\mathbf{\theta}_2}(\mathbf{x}_2 \mid \mathbf{z})$ are parameterized by neural networks with parameters $\phi$, $\mathbf{\theta}_1$ and $\mathbf{\theta}_2$ respectively. The first term in Equation \ref{eq:VCCA_loss} is the KL divergence between the approximate posterior $q_{\phi}(\mathbf{z} \mid \mathbf{x}_1)$ and the prior $p(\mathbf{z})$. The second term is the expected data log-likelihood under the approximate posterior distribution. 

We note that, DVCCA is based on the estimation of a single latent posterior distribution, $q_{\phi}(\mathbf{z} \mid \mathbf{x}_1)$. Therefore, the resulting representation  is dependent on the reference modality from which the joint latent representation is encoded, and may therefore bias the estimation of the latent representation. Finally, \citet{Wang2016a} introduce a variant of DVCCA, VCCA-private, which extracts the private, $\mathbf{h}$, in addition to shared,  $\mathbf{z}$, latent information (i.e. assuming a latent variable model illustrated in Figure \ref{fig:AElatent_c}). VCCA-private has the following objective: 

\begin{equation}\label{eq:VCCA_private_loss}
\begin{aligned}
&\mathcal{L}^{\text{VCCA}}_{\text {private }}=-D_{K L}\left(q_{\mathbf{\phi}}(\mathbf{z} \mid \mathbf{x}_1) \| p(\mathbf{z})\right)\\
&-D_{K L}\left(q_{\mathbf{\phi}}\left(\mathbf{h}_{1} \mid \mathbf{x}_1\right) \| p\left(\mathbf{h}_{1}\right)\right)-D_{K L}\left(q_{\mathbf{\phi}}\left(\mathbf{h}_{2} \mid \mathbf{x}_2\right) \| p\left(\mathbf{h}_{2}\right)\right)\\
&+\mathbb{E}_{q_{\phi}(\mathbf{z} \mid \mathbf{x}_1), q_{\phi}\left(\mathbf{h}_{1} \mid \mathbf{x}_1\right)}\left[\log p_{\mathbf{\theta}}\left(\mathbf{x}_1 \mid \mathbf{z}, \mathbf{h}_{1}\right)\right]\\
&+\mathbb{E}_{q_{\phi}(\mathbf{z} \mid \mathbf{x}_1), q_{\phi}\left(\mathbf{h}_{2} \mid \mathbf{X}_2\right)}\left[\log p_{\mathbf{\theta}}\left(\mathbf{x}_2 \mid \mathbf{z}, \mathbf{h}_{2}\right)\right]
\end{aligned}
\end{equation}

where $\textbf{h}_1$ and $\textbf{h}_2$ are the private latent variables for $\textbf{x}_1$ and $\textbf{x}_2$, respectively.

Equation \ref{eq:VCCA_private_loss} has two addition terms compared to Equation \ref{eq:VCCA_loss}; the KL divergence between the approximate posterior distribution $q_{\mathbf{\phi}}(\mathbf{h}_{1} \mid \mathbf{x}_1)$ and the prior of the private variables $p(\mathbf{h}_{1})$ and similarly for $\mathbf{h}_{2}$ and $\mathbf{x}_2$.

\subsection*{Multi-Channel Variational Autoencoder (mcVAE)}

In a similar approach to the single view VAE, mcVAE \citep{Antelmi2019} approximates the posterior distribution of modality $m$ with $q_{\varphi_m}(\mathbf{z}_m|\mathbf{x}_m)$. The mcVAE optimises the following ELBO:

\begin{equation}
\mathcal{L}^{\text{mcVAE}} = \sum_{m=1}^M\left[\mathbb{E}_{q_{\varphi_m}(\mathbf{z}_m \mid \mathbf{x}_m)}\left[\sum_{n=1}^M \log p_{\theta_n}\left(\mathbf{x}_{n} \mid \mathbf{z}_m\right)\right]-D_{K L}\left(q_{\varphi_m}(\mathbf{z}_m \mid \mathbf{x}_m) \| p(\textbf{z})\right)\right]
\end{equation}
where the first term corresponds to the negative of the same-modality and cross-modality reconstruction loss obtained by decoding from the latent representation $\mathbf{z}_m$ using the decoder network $p_{\theta_n}\left(\mathbf{x}_{n} \mid \mathbf{z}_m\right)$. The second term is a KL divergence term enforcing each modality-specific posterior, $q_{\varphi_i}(\mathbf{z}_m|\mathbf{x}_m)$, to be similar the common prior $p(\mathbf{z})$. These two terms encourage a coherence across latent representations from different modalities. 

\citet{Antelmi2019} go on to introduce sparse-mcVAE which uses variational dropout \citep{Kingma2015} to enforce parsimonious and informative latent representations. This involves sampling $\textbf{z}$ from a dropout posterior: 
\begin{equation}
    \textbf{z} \sim \mathcal{N}\left(\mu ; \alpha \mu^2\right)
\end{equation}
where $\alpha=\frac{p}{1-p}$ and $p$ is a hyperparameter known as the drop rate, and using an improper log-scale uniform prior: 
\begin{equation}
\centering
\begin{gathered}
p(\textbf{z})=\prod p\left(\left|z_i\right|\right) \\
p(\ln |z_i|)=\text {const} \Leftrightarrow p(|z_i|) \propto \frac{1}{|z_i|}
\end{gathered}  
\end{equation}
where $z_i$ is a factor of $\textbf{z}$. Both mcVAE and sparse-mcVAE are implemented in the \texttt{multi-view-AE} library.

\subsection*{Multi-modal Variational Autoencoder (MVAE) and multi-elbo MVAE}

\citet{Wu2018} introduced a multi-modal Variational Autoencoder (MVAE) which uses a Product-of-Experts (PoE) approach to approximate a joint posterior distribution. MVAE optimises the following ELBO:

\begin{equation}\label{eq:mvae}
\mathcal{L}^{\text{MVAE}} = \mathbb{E}_{q_{\phi}(\textbf{z} \mid \textbf{X})}\left[\sum_{m=1}^{M} \log p_{\theta}\left(\textbf{x}_{m} \mid \textbf{z}\right)\right]-D_{K L}\left(q_{\phi}(\textbf{z} \mid \textbf{X})\| p(\textbf{z})\right).
\end{equation} 

Here, $q_{\phi}(\textbf{z} \mid \textbf{X})$ is the joint posterior given by $q_{\phi}(\textbf{z} \mid \textbf{X}) \propto p(\textbf{z}) \prod_{m=1}^M q_{\phi_m}\left(\textbf{z} \mid \textbf{x}_{m}\right)$ where $q_{\phi_m}\left(\textbf{z} \mid \textbf{x}_{m}\right)$ is the encoder network for the modality $m$. Assuming each encoder network follows a gaussian distribution such that $q\left(\textbf{z} \mid \textbf{x}_{m}\right)=\mathcal{N}(\mu_m, \sigma_m)$, the parameters of joint posterior distribution can be computed using the Inverse-Variance Weighted (IVW) method \citep{Hwang2021}: 
\begin{equation*}
  \mu \triangleq \frac{\sum_{m=1}^{M} \mu_{m} / \sigma_{m}^{2}}{\sum_{m=1}^{V} 1 / \sigma_{m}^{2}} \quad \text { and } \quad \sigma^{2} \triangleq \frac{1}{\sum_{m=1}^{M} 1 / \sigma_{m}^{2}}.  
\end{equation*}

However, Equation \ref{eq:mvae} does not train the individual encoders and decoders to deal with missing data at test time. As such, \citet{Wu2018} introduce a sub-sampling paradigm which includes the ELBO term for the whole dataset, the ELBO terms using a single modality, and $k$ ELBO terms using $k$ randomly chosen subsets, $\textbf{X}_k$. The sub-sampled objective is given by:

\begin{equation}
   \operatorname{ELBO}\left(\textbf{x}_1, \ldots, \textbf{x}_M\right)+\sum_{m=1}^M \operatorname{ELBO}\left(\textbf{x}_m\right)+\sum_{j=1}^k \operatorname{ELBO}\left(\textbf{X}_j\right). 
\end{equation}
In our implementation, we chose to exclude the ELBO terms using random subsets as these were not used in subsequent related work \citep{Sutter2021,Hwang2021} and some of analysis conducted by the authors of the original paper\footnote{https://github.com/mhw32/multimodal-vae-public}. We term this variant of the MVAE, me\_mVAE.

\subsection*{Variational Mixture-of-Experts Autoencoder (mmVAE)}

However, overconfident but miscalibrated experts, i.e. the individual posterior distributions, may bias the joint posterior distribution in the PoE approach \citep{Shi2019}. \citet{Shi2019} attempt to mitigate this problem by combining latent representations across modalities using a Mixture-of-Experts (MoE) approach and thus taking a vote amongst experts. \citet{Shi2019} assume the joint posterior is given by $q_{\Phi}\left(\textbf{z} \mid \textbf{X}\right)=\sum_{i=1}^{M} \frac{1}{M}q_{\phi_{m}}\left(\textbf{z} \mid \textbf{x}_{m}\right)$. mmVAE optimises the following objective, which uses a k-samples Importance Weighted Autoencoder (IWAE) lower bound \citep{Burda2016}:

\begin{equation}
\mathcal{L}_{\mathrm{IWAE}}^{\mathrm{mmVAE}}=\frac{1}{M} \sum_{m=1}^{M} \mathbb{E}_{\textbf{z}_{m}^{1: K} \sim q_{\phi_{m}}\left(\textbf{z} \mid \textbf{x}_{m}\right)}\left[\log \frac{1}{K} \sum_{k=1}^{K} \frac{p(\textbf{z}) \prod_{m=1}^M p_{\theta_m}\left(\textbf{x}_m \mid \textbf{z}^{k}\right)}{q_{\Phi}\left(\textbf{z}^{k} \mid \textbf{X}\right)}\right]
\end{equation}
where $k$ is the number of samples drawn from the posterior distribution, $q_{\phi_{m}}\left(\textbf{z} \mid \textbf{x}_{m}\right)$, for each data point. However, mmVAE only takes each uni-modal posterior separately into account during training and thus does not explicitly aggregate information from multiple modalities in the latent representation. Furthermore, it has been found that the sub-sampling paradigm used in mixture-based multi-modal VAEs enforces an upper bound on the multi-modal ELBO thus preventing a tight approximation of the joint distribution \citep{Daunhawer2021}.

\subsection*{Multi-View Total Correlation Autoencoder (MVTCAE)}

There have been a few approaches to tackle the pitfalls of Product-of-Experts (PoE) and Mixture-of-Experts (MoE) multi-modal VAEs. With the Multi-View Total Correlation Autoencoder (MVTCAE), \citet{Hwang2021} approach the variational inference problem from a total correlation (TC) perspective where TC is defined as the KL divergence of the joint distribution from the factored marginal distributions \citep{Watanabe1960}:

\begin{equation}
    T C(\textbf{X}) \triangleq D_{K L}\left(p_{\text{data}}(\textbf{X}) \| \prod_{m=1}^{M} p_{\text{data}}\left(\textbf{x}_{m}\right)\right)
\end{equation}

where $p_{\text{data}}(\textbf{X})$ is the true data distribution. The aim, is then to find the encoder $q_{\phi}(\textbf{z} \mid \textbf{x})$ such that the knowledge of $\textbf{z}$ would reduce TC as much as possible and is formulated in the following objective:

\begin{equation}
    T C_{\theta}(\textbf{X} ; \textbf{z}) \triangleq T C(\textbf{X})-T C_{\theta}(\textbf{X} \mid \textbf{z})
\end{equation}
with the conditional TC is given by: 

\begin{equation}
T C_{\theta}(\textbf{X} \mid \textbf{z}) \triangleq \mathbb{E}_{p_{\theta}(\textbf{z})}\left[D_{K L}\left(p_{\theta}(\textbf{X} \mid \textbf{z}) \| \prod_{m=1}^{M} p_{\theta}\left(\textbf{x}_{m} \mid \textbf{z}\right)\right)\right].
\end{equation}
\citet{Hwang2021} show that the TC objective can be solved by optimising the following variational lower bound: 

\begin{equation}
\begin{gathered}
\mathcal{L}^{\mathrm{MVTCAE}}= \frac{M-\alpha}{M} \sum_{m=1}^{M}\left[\mathbb{E}_{p(\textbf{z} \mid \textbf{x})}\left[\log p_{\theta_{m}}\left(\textbf{x}_{m} \mid \textbf{z}\right)\right]\right] -\\
\beta\left[(1-\alpha)D_{K L}\left(q_{\phi}(\textbf{z} \mid \textbf{X}) \| p\left(\textbf{z}\right)\right) + 
\frac{\alpha}{M} \sum_{m=1}^{M} D_{K L}\left(q_{\phi}(\textbf{z} \mid \textbf{X}) \| q_{\phi_{m}}\left(\textbf{z} \mid \textbf{x}_{m}\right)\right)\right]
\end{gathered}
\end{equation}
where $\beta$ weights the regularisation and $\alpha$ is used to weight the different divergence terms. The KL terms between the joint representation $q_{\phi}(\textbf{z} \mid \textbf{X})$ and the individual encoders $q_{\phi_{m}}\left(\textbf{z} \mid \textbf{x}_{m}\right)$ (known as Conditional Variational Information Bottlenecks, CVIB) help calibrate each encoder to the joint encoder, encouraging the joint representation to encode information from all modalities. MVTCAE uses a PoE joint posterior distribution given by; $q(\textbf{z} \mid \textbf{X}) \propto \prod_{m=1}^M q\left(\textbf{z} \mid \textbf{x}_{m}\right)$.

\subsection*{Mixture-of-Products-of-Experts VAE (MoPoEVAE)}

\citet{Sutter2021} instead opt to model the joint encoding distribution by a Mixture-of-Product-of-Experts (MoPoE). Here they define subset joint posteriors given by the product of each expert in the subset, $\textbf{X}_k$: 

\begin{equation}
q_\phi\left(\textbf{z} \mid \textbf{X}_k\right) \propto \prod_{j=1}^{k} q_{\phi_j}\left(\textbf{z} \mid \textbf{x}_j\right)
\end{equation}
for each subset of the powerset $\mathcal{P}(\textbf{X})$. The joint posterior is given by a mixture of the subset joint posteriors, $q_\phi\left(\textbf{z} \mid \textbf{X}_k\right)$:

\begin{equation}
q_\phi(\textbf{z} \mid \textbf{X})=\frac{1}{N} \sum_{\textbf{X}_k \in \mathcal{P}(\textbf{X})} q_\phi\left(\textbf{z} \mid \textbf{X}_k\right)
\end{equation}
where $N$ is the total number of subsets. The MoPoE objective is then defined as: 

\begin{equation}
\mathcal{L}^{\text{MoPoE}} =\mathbb{E}_{q_\phi(\textbf{z} \mid \textbf{X})} \left[\log p_\theta\left(\textbf{X} \mid \textbf{z}\right)\right]-D_{\mathrm{KL}}\left(\frac{1}{N} \sum_{\textbf{X}_k \in \mathcal{P}(\textbf{X})} q_\phi\left(\textbf{z} \mid \textbf{X}_k\right) \| p_\theta(\textbf{z})\right).
\end{equation}

The MoPoE objective shares similarities with the MVAE sub-sampled objective, with a few key distinctions \citep{senellart2023}. In the MVAE sub-sampled objective, the ELBO is computed for all subsets for every data point. In contrast, the MoPoE objective involves sampling a single subset for each data point at each iteration, computing the ELBO for only that specific subset. Additionally, unlike the MVAE objective, which considers only the subset of all modalities, the uni-modal subsets, and $K$ other random subsets, the MoPoE objective includes all possible subsets of $\textbf{X}$, which can be computationally costly for a large number of modalities. 

\subsection*{Generalised Product-of-Experts Variational Autoencoder (weighted mVAE)}

Another approach to dealing with the problem of the Product-of-Experts is to use the generalised Product-of-Experts (gPoE) \citep{Cao2014} approach as done in the weighted mVAE \citep{Lawryaguila2023}. Here, the joint posterior distribution is given by: 

\begin{equation}\label{eq:gPoE}
q_{\phi}\left(\textbf{z} \mid \textbf{X}\right) = \frac{1}{K} \prod_{m=1}^{M} q_{\phi_{m}}^{\alpha_{m}}\left(\textbf{z} \mid\textbf{x}_{m}\right) 
\end{equation} 

where $\alpha_{m}$ is a weighting for modality $m$ learnt during training. Similarly to the PoE approach, we can compute the parameters of the joint posterior distribution:
\begin{equation}
\boldsymbol{\mu} = \frac{\sum_{m=1}^{M} \boldsymbol{\mu}_{m}\boldsymbol{\alpha}_{m} / \boldsymbol{\sigma}_{m}^{2}}{\sum_{m=1}^{M} \boldsymbol{\alpha}_{m} / \boldsymbol{\sigma}_{m}^{2}} \quad \text { and } \quad \boldsymbol{\sigma}^{2} = \sum_{m=1}^{M} \frac{1}{ \boldsymbol{\alpha}_{m} / \boldsymbol{\sigma}_{m}^{2}}.
\end{equation}

The objective function is then the same as that given in Equation \ref{eq:mvae}. The gPoE can be thought of as a middle ground between the PoE and MoE pooling operators. MoE is not very sensitive to consensus across modalities and will give lower probability to regions where experts are in agreement than PoE. However, PoE is highly sensitive to overconfident miscalibrated experts which can result in suboptimal latent space and data reconstruction. The gPoE strikes a balance, being more sensitive to consensus across modalities than MoE, while avoiding the problems of overconfident experts by down-weighing them during training through the weighting term.

\subsection*{Multi-modal Jensen-Shannon divergence model (mmJSD)}

For many of the models discussed thus far, the objective functions are designed such that only either the uni-modal or joint posterior distributions are optimised. To address this, \citet{Sutter2021b} introduce another generative model, mmJSD, which uses a Jensen-Shannon divergence in the objective function to optimised both the uni-modal and joint posterior distributions. The Jensen-Shannon divergence is defined as a sum of KL divergences between $M+1$ probability distributions $q_m(\textbf{z})$ and their mixture distribution:

\begin{equation}
J S_{\boldsymbol{\pi}}^{M+1}\left(\left\{q_m(\textbf{z})\right\}_{m=1}^{M+1}\right)=\sum_{m=1}^{M+1} \pi_m D_{KL}\left(q_{\phi_m}(\textbf{z}) \| \sum_{v=1}^{M+1}\pi_v q_{\phi_\nu}(\textbf{z})\right).
\end{equation}

where $\pi_m$ and $\pi_v$ are the weights for distributions $m$ and $v$, respectively. mmJSD defines a new ELBO which uses the Jensen-Shannon divergence: 
\begin{equation}\label{eq:mmJSD}
L = E_{q_\phi(\textbf{z} \mid \textbf{X})}\left[\log p_\theta(\textbf{X} \mid \textbf{z})\right]-J S_{\boldsymbol{\pi}}^{M+1}\left(\left\{q_{\phi_{m}}\left(\textbf{z} \mid \textbf{x}_m\right)\right\}_{m=1}^M, p_\theta(\textbf{z})\right).
\end{equation}
The Jensen-Shannon divergence term can be re-written by considering a "dynamic-prior" defined as a function $f$ of the uni-modal posterior distributions $q_{\phi_{m}}\left(\textbf{z} \mid \textbf{x}_m\right)$ and the prior distribution $p_\theta(\textbf{z})$:
\begin{equation}
    p_f(\textbf{z} \mid \textbf{X})=f\left(\left\{q_{\phi_\nu}\left(\textbf{z} \mid \textbf{x}_\nu\right)\right\}_{\nu=1}^M, p_\theta(\textbf{z})\right).
\end{equation}

The "dynamic prior" is not technically a prior distribution as it does not reflect prior knowledge of the data. It does, however, incorporate the prior knowledge that all modalities share common factors \citep{Sutter2021b}. Assuming the chosen "dynamic prior" is well-defined, as the authors prove for MoE and PoE "dynamic priors", Equation \ref{eq:mmJSD} becomes:
\begin{equation}
\begin{aligned}\mathcal{L}^{\text{mmJSD}} &= E_{q_\phi(\textbf{z} \mid \textbf{X})}\left[\log p_\theta(\textbf{X} \mid \textbf{z})\right]-\sum_{m=1}^M \pi_m D_{KL}\left(q_{\phi_m}\left(\textbf{z} \mid \textbf{x}_m\right) \| p_f(\textbf{z} \mid \textbf{X})\right) \\ & -\pi_{M+1} D_{KL}\left(p_\theta(\textbf{z}) \| p_f(\textbf{z} \mid \textbf{X})\right).
\end{aligned}
\end{equation}
In the \texttt{multi-view-AE} package we implement mmJSD with a PoE "dynamic prior". This choice aligns with the implementations used in previous work that employed mmJSD for comparison purposes \citep{Hwang2021}.

\subsection*{Mixture-of-experts multi-modal VAE Plus (MMVAE+)}

As highlighted by \citet{Daunhawer2021}, current multi-modal VAEs struggle to match the generative quality of uni-modal models, particularly if there is a high degree of modality-specific variation. One way to address the poor generative quality of multi-modal VAEs is by incorporating modality-specific private latents to explicitly encode modality-specific variation in addition to variation shared across modalities. \citet{Palumbo2022} introduce MMVAE+, a variant of the MoE multi-modal VAE (mmVAE) with modality-specific latents, $\textbf{h}$, which assumes the latent variable model show in Figure \ref{fig:AElatent_c}. The joint posterior is given by a Mixture-of-Experts of the form; $q_{\Phi}\left(\textbf{z}, \textbf{h} \mid \textbf{X}\right)=\frac{1}{M} \sum_{m=1}^Mq_{\phi_{\textbf{z}_m}}\left(\textbf{z} \mid \textbf{x}_m\right) q_{\phi_{\textbf{h}_m}}\left(\textbf{h}_m \mid \textbf{x}_m\right)$ where $q_{\phi_{\textbf{z}_m}}\left(\textbf{z} \mid \textbf{x}_m\right)$ is the shared encoder for modality $m$ and $q_{\phi_{\textbf{h}_m}}\left(\textbf{h}_m \mid \textbf{x}_m\right)$ is the private modality-specific encoder.

The MMVAE+ objective is given by:

\begin{equation}
\begin{aligned}
     &\mathcal{L}^{\mathrm{MMVAE}+}=
     \\& \frac{1}{M} \sum_{m=1}^M \mathbb{E}_{\substack{q_{\phi_{\textbf{z}_m}\left(\textbf{z} \mid \textbf{x}_m\right)} \\ q_{\phi_{\textbf{h}_m}\left(\textbf{h}_m \mid \textbf{x}_m\right)}}}\left[ \log \left(
    \frac{p_{\theta_m}\left(\textbf{x}_m \mid \textbf{z}, \textbf{h}_m\right)p\left(\textbf{z}\right) p\left(\textbf{h}_m\right)}{q_{\phi_z}\left(\textbf{z}\mid \textbf{X}\right) q_{\phi_{\textbf{h}_m}}\left(\textbf{h}_m \mid \textbf{x}_m\right)} \prod_{n \neq m} \mathbb{E}_{p_{\psi_n}\left(\tilde{\textbf{h}}_n\right)}\left[p_{\theta_n}\left(\textbf{x}_n \mid \textbf{z}, \tilde{\textbf{h}}_n\right)\right] 
    \right)
    \right]   
\end{aligned}
\end{equation}

where $p_{\psi_1}\left(\tilde{\textbf{h}}_1\right), \ldots, p_{\psi_M}\left(\tilde{\textbf{h}}_M\right)$ are modality specific auxilary priors with parameterised variance. The modality specific information, $\textbf{h}_m$, is only used for same-modality reconstruction. For cross-modality reconstruction, the private information is drawn from the respective prior, $\tilde{\textbf{h}}_m \sim p_{\psi_m}\left(\tilde{\textbf{h}}_m\right)$. 


\subsection*{Disentangled multi-modal Variational Autoencoder (DMVAE)}

Similarly, \citet{Lee2021} introduce a shared and private multi-modal VAE, which they term disentangled multi-modal
variational autoencoder (DMVAE). DMVAE is similar to MMVAE+ in that it considers a set of modality specific latents $\textbf{h}$ as well as the shared latents $\textbf{z}$. In contrast to MMVAE+, DMVAE defines the joint posterior distribution as a Product-of-Experts:

\begin{equation*}
    q\left(\textbf{z} \mid \textbf{X}\right) \propto p\left(\textbf{z}\right) \prod_{i=1}^M q\left(\textbf{z} \mid \textbf{x}\right).
\end{equation*}

The DMVAE objective is given by:

\begin{equation}
\begin{aligned}
& \mathcal{L}^{\text{DMVAE}} = \sum_{m=1}^M \left(\lambda_m \mathbb{E}_{q_{\phi_{\textbf{h}_m}}\left(\textbf{h}_m \mid \textbf{x}_m\right),\, q_\phi\left(\textbf{z} \mid \textbf{X}\right)}\left[\log p_\theta\left(\textbf{x}_m\mid \textbf{h}_m, \textbf{z} \right)\right]\right. \\ & - D_{\mathrm{KL}}\left(q_{\phi_{\textbf{h}_m}}\left(\textbf{h}_m \mid \textbf{x}_m\right)|| p\left(\textbf{h}_m\right)\right)- D_{\mathrm{KL}}\left(q_\phi\left(\textbf{z} \mid \textbf{X}\right) \| p\left(\textbf{z}\right)\right) \\ & + \sum_{n=1}^N \left(\lambda_m \mathbb{E}_{q_{\phi_{\textbf{h}_m}}\left(\textbf{h}_m \mid \textbf{x}_m\right),\, q_{\phi_{\textbf{z}_n}}\left(\textbf{z} \mid \textbf{x}_n\right)}\left[\log p_\theta\left(\textbf{x}_m \mid \textbf{h}_m, \textbf{z}\right)\right]\right. \\ & \left.\left.- D_{\mathrm{KL}}\left(q_{\phi_{\textbf{h}_m}}\left(\textbf{h}_m\mid \textbf{x}_m\right)|| p\left(\textbf{h}_m\right)\right)-D_{\mathrm{KL}}\left(q_{\phi_{\textbf{z}_n}}\left(\textbf{z} \mid \textbf{x}_n\right)|| p\left(\textbf{z}\right)\right)\right)\right)
\end{aligned}
\end{equation}
where $\lambda_m$ balances the reconstruction from different modalities. The first term in the objective models the accuracy of the reconstruction using the modality specific private representation and the shared latent representation generated by the joint posterior distribution, $q_\phi(\textbf{z}|\textbf{X})$. The second set of reconstruction terms, correspond to the reconstruction term obtained by cross-modality, $n\neq m$, reconstruction and same-modality, $n=m$, reconstruction using the shared representation generated from modality $n$. The reconstruction terms are compensated by a number of KL terms which encourage the relevant private and shared representation to be similar to their respective priors. 

\subsection*{Multi-view Adversarial Autoencoder (mAAE)}

As with VAEs, AAEs can be extended to multiple modalities or views. An AAE consists of encoder and decoder networks, $f_e$ and $f_d$, and a neural network discriminative model, $D(\textbf{z})$, which computes the probability that a sample $\textbf{z}$ was drawn from the prior $p(\textbf{z})$ or from the generative model $G(\textbf{x})$, the encoding function, $f_e(\textbf{x})$. In the \texttt{multi-view-AE} package, we implement a multi-modal AAE with the following objective function:

\begin{equation}
\mathcal{L}^{\text{mAAE}}=L_{\text{GAN}}+L_{\text{MSE}}
\end{equation}

where $L_{\text{GAN}}$ is a min-max loss given by:

 \begin{equation}\label{eq:multi-view_Adversarial}
L_{\text{GAN}} =\frac{1}{M} \sum_{m=1}^M\min _{G_m} \max_{D} \mathbb{E}_{\textbf{z} \sim p(\textbf{z})}[\log D(p(\textbf{z}))]+\mathbb{E}_{\textbf{x}_m \sim p(\textbf{x}_m)}[\log (1-D(G_m(\textbf{x}_m))]
\end{equation}

and $L_{\text{MSE}}$ is the reconstruction loss for same-modality and cross-modality reconstruction: 

\begin{equation}
L_{\text{MSE}} = \frac{1}{M} \sum_{m=1}^M \frac{1}{M} \sum_{n=1}^M \|\textbf{x}_m-f_d^m(f_e^n(\textbf{x}_n))\|_2^2
\end{equation}
Our implementation is closely related to the AdvCAE model \citep{Wang2019b} with a scaled-back objective function. Whilst our implementation does not directly recreate previous work, we hope that it acts as a building block for developers and researchers wishing to use multi-modal AAE models. 

\subsection*{Multi-view Adversarial Autoencoder with wasserstein loss (mWAE)}

To deal with instability with GAN (Generative Adversarial Network) training \citep{Arjovsky2017}, we implement a multi-modal AAE with a wasserstein loss based on the wasserstein GAN \citep{Arjovsky2017}. Instead of a discriminator network to classify samples as "real" or "fake", the wasserstein GAN uses a critic network, $C(\textbf{z})$, to score the realness or fakeness of a sample. The wasserstein GAN uses a wasserstein distance, a metric to measure the distance between two probability distributions, to encourage the output of the generator to be close to the prior distribution. The mWAE objective function is given by:

\begin{equation}\label{eq:mWAE}
\mathcal{L}^{\text{mWAE}}=L_{\text{MSE}} - (L_{\text{Critic}}+L_{\text{Generator}})
\end{equation}
where:
\begin{equation}\label{eq:critic_generator}
\begin{aligned}
L_{\text{Critic}} &= \frac{1}{M} \sum_{m=1}^M\max_{C} \mathbb{E}_{\textbf{z} \sim p(\textbf{z})}[C(p(\textbf{z}))]-\mathbb{E}_{\textbf{x}_m \sim p(\textbf{x}_m)}[C(G(\textbf{x}_m))] \\
L_{\text{Generator}} &= \frac{1}{M} \sum_{m=1}^M\max_{G_m}\mathbb{E}_{\textbf{x}_m \sim p(\textbf{x}_m)}[C(G_m(\textbf{x}_m))].
\end{aligned}
\end{equation}
As with the mAAE, our wAAE implementation does not directly recreate previous work.

\section{Software architecture}

\begin{figure}
     \centering
         \includegraphics[width=\textwidth, scale=0.6]{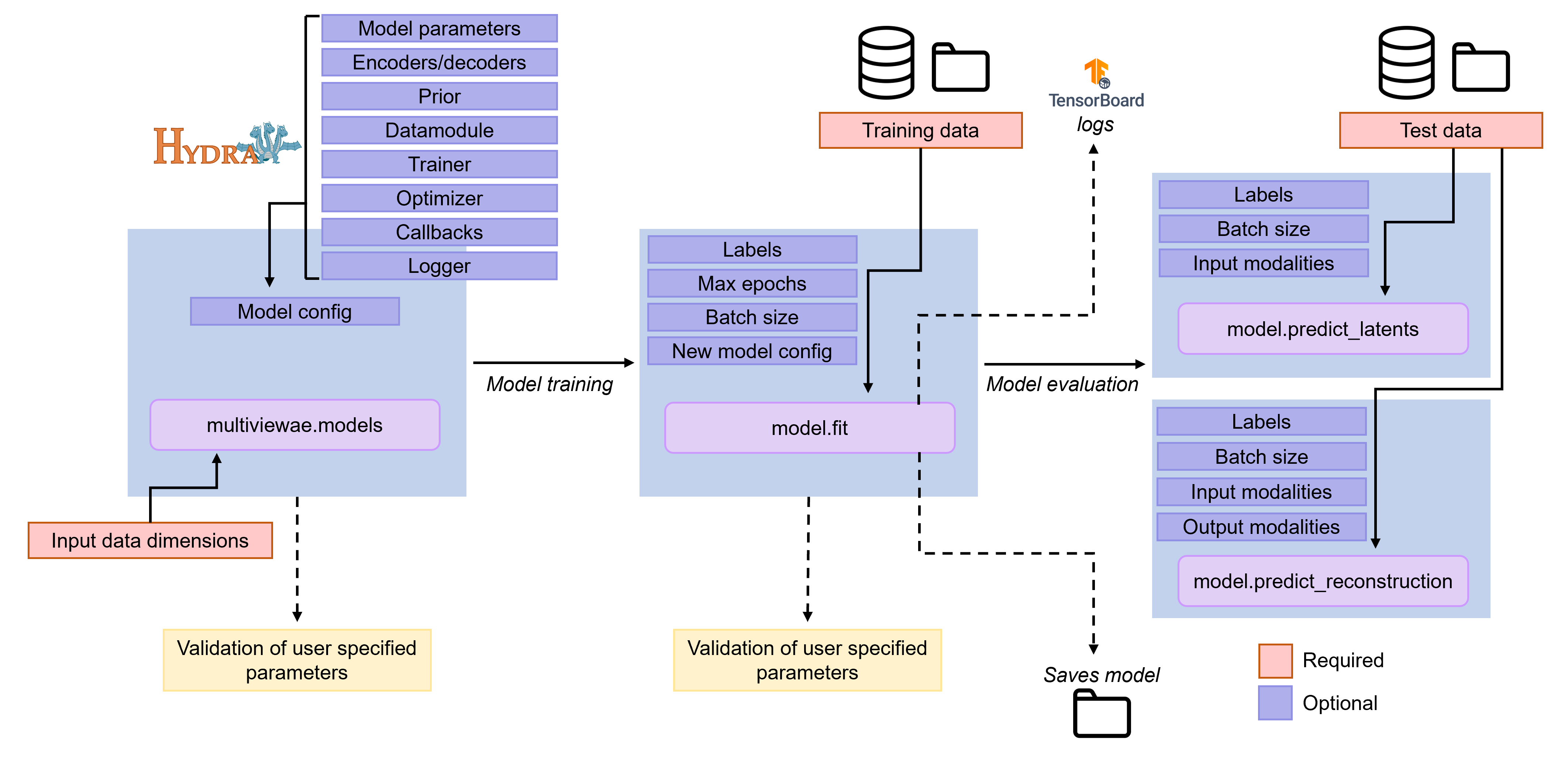}
        \caption{Schematic diagram of the \texttt{multi-view-AE} package.}\label{fig:lib_schematic}
\end{figure}

\begin{figure}
     \centering
         \includegraphics[width=\textwidth, scale=0.6]{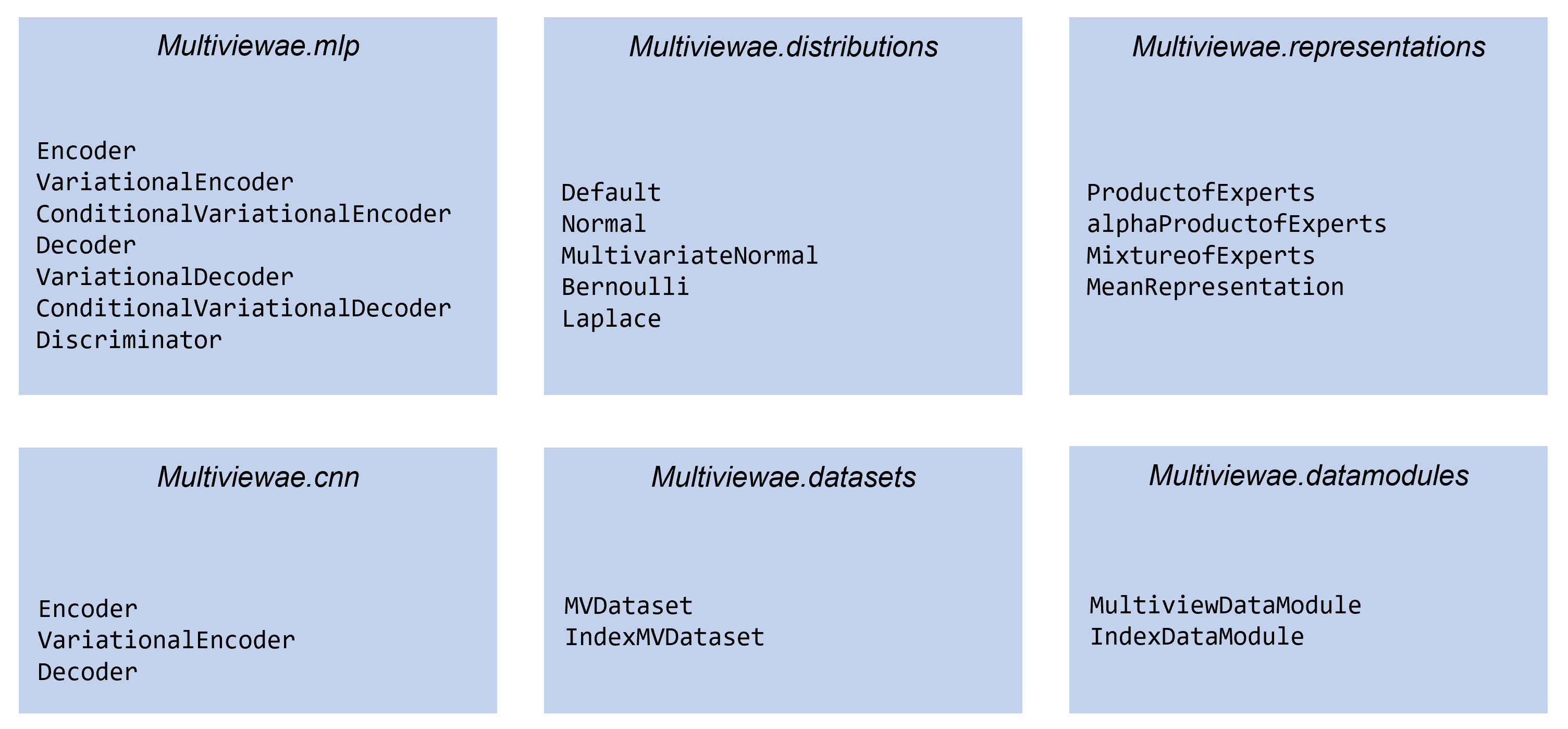}
        \caption{Implemented classes grouped by broad category implemented in \texttt{multi-view-AE} in addition to the model classes.}\label{fig:implemented_classes}
\end{figure}

All models in the \texttt{multi-view-AE} package are implemented using a common API for ease of use and in such a way that models can be trained in a few lines of code. To train a multi-view autoencoder model with \texttt{multi-view-AE}, first a model object is initialised with the relevant parameters in an easy-to-configure yaml file. Next, the model is trained with the \texttt{fit()} method using the specified data. Following fitting, the saved model object can be used for further analysis: predicting the latent variables, using \texttt{predict\_latent()}, or data reconstructions, using \texttt{predict\_reconstruction()}. These three functions are implemented within the \texttt{BaseModelAE}, \texttt{BaseModelVAE} or \texttt{BaseModelAAE} class depending on whether the model is a vanilla, adversarial, or variational model. These are abstract classes, that cannot be instantiated directly, which inherit from \texttt{Pytorch-lightning} (the latter two classes also inheriting from \texttt{BaseModelAE}). As a minimum, each model child class implements its own \texttt{encode()}, \texttt{decode()}, and \texttt{loss\_function()} methods which are called during model training and evaluation. Each model may then also have additional model specific methods depending, for example, on the training framework or method of application to new data. All models are implemented in \texttt{PyTorch} using the \texttt{PyTorch-Lightning} wrapper. A schematic of the \texttt{multi-view-AE} package is shown in Figure \ref{fig:lib_schematic}. Figure \ref{fig:implemented_classes} shows the groups of classes accompanying the model implementations in the \texttt{multi-view-AE} package. More detail on these classes, which can be specified in the user provided configuration file, is provided in the following paragraphs.

\textbf{Network architectures.} The \texttt{multi-view-AE} package contains built in neural-network classes. These are organised into \texttt{Encoder} and \texttt{Decoder} classes which each implement a \texttt{forward()} method to carry out a forward pass through the network layers. \texttt{VariationalEncoder} and \texttt{VariationalDecoder} classes inherit from their non-variational counterparts. There are both multilayer perceptron (MLP) and convolutional neural network (CNN) implementations designed in such a way as to maximise flexibility and ease of use. There are also conditional variants of the MLP networks for implementing conditional Variational Autoencoders (cVAE).

\textbf{Distributions.} There are a number of distributions implemented in \texttt{multi-view-AE} that are called as required. These include; \texttt{Normal}, \texttt{MultivariateNormal}, \texttt{Bernoulli}, \texttt{Laplace} and \texttt{Default}. Each class implements \texttt{\_sample()} and \texttt{log\_likelihood()} methods, as minimum, for sampling from the distribution and calculating the log likelihood, respectively. Where applicable, the classes inherit from the relevant  \texttt{PyTorch} class. The \texttt{Default} class is an artificial distribution designed for data with unspecified distribution. It is used so that \texttt{log\_likelihood()}, which in this instance computes the MSE between the input data and the reconstruction, and \texttt{\_sample()} methods can be called by the model class like for other distribution classes. 

\textbf{Representation functions.} The \texttt{ProductOfExperts}, \texttt{alphaProductOfExperts}, \texttt{MixtureOfExperts}, and \texttt{MeanRepresentation} functions are used to model the joint representation from the separate representations of each modality. Each class implements a \texttt{forward()} method. The choice of representation function is inherence to the implemented model, and as such does not require editing in the configuration file provided by the user.  

\textbf{Datasets and data modules.} \texttt{multi-view-AE} has two datamodule classes; \texttt{MultiviewDataModule} and \texttt{IndexDataModule}. These classes, which inherit from \texttt{PytorchLightning.DataModule}, encapsulate all the steps for processing the data. To load the data, they call dataset classes \texttt{MVDataset} and \texttt{IndexMVDataset} depending on whether the input is the data matrix, or a list of indices and path variable to load the data. 

\textbf{Trainers, optimizers and logging.} In addition to the \texttt{multi-view-AE} implementations described above, the user can change the parameters of the trainer and logger used during the training procedure. Models within the \texttt{multi-view-AE} package are trained by default using a \texttt{PyTorchLightning.Trainer} and a \texttt{TensorBoard} logger to monitor training. Whilst the user is free to use other loggers, e.g. \texttt{wandb}, these have not been tested within the package. The user can also chose the optimizer algorithm used for optimisation from the \texttt{torch.optim} package. However, note that only the Adam optimizer has been tested within the \texttt{multi-view-AE} package. 

\section{Benchmarking}

To illustrate the efficacy of our implementions, we validated some of the implemented models by reproducing key results from previous papers. One of the experiments presented in the paper was reproduced using the \texttt{multi-view-AE} implementations using the same network architectures, modelling choices, and training parameters. The code to reproduce the benchmarking experiments is available at the \texttt{multi-view-AE} GitHub page.

\subsection{Datasets and parameter settings}\label{Sec:benchmarking_datasets}

We use the PolyMNIST and BinaryMNIST datasets to assess the performance of our \texttt{multi-view-AE} implementations. The PolyMNIST dataset \citep{Sutter2021} is composed of MNIST images from 5 different modalities, each modality exhibiting a different background and writing style but sharing the same digit label. As such, the digit label is a common factor of variation across modalities whilst the background and writing style are modality specific. There are 60,000 training samples and 10,000 test samples. The BinaryMNIST \citep{lecun2010} dataset consists of the MNIST images with a black background and the digit in white text. To make this a multi-modal dataset, one modality is composed of images, whilst another consists of one-hot-encoded digit labels. 

\subsubsection{PolyMNIST experiments}

For mmVAE, MoPoEVAE, mmJSD, MVTCAE we reproduce the results from \citet{Hwang2021} for the PolyMNIST dataset. We use previously used CNN encoder and decoder architectures \citep{Sutter2021}, a latent dimension of 512, Gaussian priors and posteriors, and Laplace likelihoods with scale $\sigma=0.75$. We train our models for 300 epochs with a batch size of 256. For MMVAE+ we use the RNN network architecture used by \citet{Palumbo2022}, 64 latent dimensions (split equally into private and shared factors), and Laplace priors, likelihoods and posteriors. We train the MMVAE+ model for 150 epochs using a batch size of 64. For all experiments, bar MMVAE+, we use $\beta=2.5$. Other model specific parameter settings can be found in the benchmarking configuration files available in the \texttt{multi-view-AE} GitHub repository. We train our models using 5 random seeds, the same seeds as used in previous work \citep{Hwang2021}.

\subsubsection{BinaryMNIST experiments}

For me\_mVAE and JMVAE we reproduce the results from \citet{Wu2018} and \citet{Suzuki2016} respectively using the BinaryMNIST dataset. We used MLP encoder and decoder networks, a latent dimension of 64, Gaussian priors and posteriors, and Categorical and Bernoulli likelihoods for the digit labels and images respectively. We train our models for 500 epochs, using the same 5 random seeds as used for the PolyMNIST experiments. 

\subsection{Evaluation metrics}

To evaluate our model performance, we use the following previously proposed evaluation metrics.

\textbf{Conditional coherence accuracy.} We follow the method implemented by \citet{Hwang2021} to measure the conditional coherence accuracy of the test set ($\uparrow$). We extract the latent representations of every subset of modalities and use them to generate modalities that are absent from the subset. The generated modalities are fed through a modality specific CNN-classifier, pre-trained on the original uni-modal training set. The accuracy of correctly matched subset labels is averaged across all subsets of the same size.

\textbf{Joint log-likelihood.} For an unseen test set, we calculate the data log-likelihood  ($\uparrow$), $\text{log} \, p(\textbf{X})$, for all modalities using $K$ samples from the joint posterior distribution. This approach was previously used by \citet{Suzuki2016} and \citet{Wu2018}.

\subsection{Results}
Table \ref{tab:benchmarking_results} shows a summary of the results of the benchmarking experiments using the \texttt{multi-view-AE} package compared to previous implementations. We can see that the \texttt{multi-view-AE} implementations perform equally well or better than previous implementations. There are a number of improvements to the \texttt{multi-view-AE} models, such as improved implementation of the Product-of-Experts function and better choice of optimizers, which could contribute to their improved performance compared to previous results.

\begin{table}
\resizebox{\textwidth}{!}{
\begin{tabular}{|l|l|l|l|l|l|}
    \hline
    \textbf{Model} & \textbf{Experiment} & \textbf{Metric} & \textbf{Paper}  & \textbf{Paper results}  & \textbf{\texttt{multi-view-AE} results} \\
    \hline
    JMVAE & BinaryMNIST &  Joint log-likelihood &   \citep{Suzuki2016} & -86.86 & -86.76$\pm$0.06\\
    \hline
    me\_mVAE & BinaryMNIST &  Joint log-likelihood &   \citep{Wu2018} & -86.26 & -86.31$\pm$0.08\\
    \hline
    MoPoEVAE & PolyMNIST &  Coherence accuracy &   \citep{Hwang2021} & 63/75/79/81 & 68/79/83/84\\
    \hline
    mmJSD & PolyMNIST  &  Coherence accuracy &   \citep{Hwang2021} & 69/57/64/67
 & 75/74/78/80\\
    \hline
    mmVAE & PolyMNIST  &  Coherence accuracy &   \citep{Hwang2021} & 71/71/71/71 & 71/71/71/71\\
    \hline
    MVTCAE & PolyMNIST  &  Coherence accuracy &   \citep{Hwang2021} & 59/77/83/86
 & 64/81/87/90\\
    \hline    
    MMVAE+ & PolyMNIST  &  Coherence accuracy &   \citep{Palumbo2022} & 85.2 & 86.6$\pm$0.07\\
    \hline    
\end{tabular}}
\caption{Benchmarking results. For each model, we replicate an experiment from a previous paper.}
\label{tab:benchmarking_results} 
\end{table}

\section{Package usage}
\subsection*{Code availability}
The code for \texttt{multi-view-AE} is available in the GitHub repository\footnote{https://github.com/alawryaguila/multi-view-AE}. Releases are tagged on GitHub and published on PyPi\footnote{https://pypi.org/project/multiviewae/}. \texttt{multi-view-AE} is compatible and tested with Python versions 3.7, 3.8, and 3.9. 

\subsection*{Example code}
Example tutorials are available in the GitHub repository in the form of Python scripts. We use the BinaryMNIST dataset, as described in Section \ref{Sec:benchmarking_datasets}, to illustrate the functionality of the \texttt{multi-view-AE} package. An excerpt of one of Python scripts is shown below. 

\begin{minted}
[frame=lines,
framesep=2mm,
baselinestretch=1.2,
bgcolor=LightGray,
fontsize=\footnotesize,
linenos
]
{python}
import torch.nn.functional as F
from torchvision import datasets
import torch
from multiviewae.models import mmVAE
import matplotlib.pyplot as plt

#Load the MNIST data
MNIST_1 = datasets.MNIST('./data/MNIST', train=True, download=True)
data_img = torch.Tensor(MNIST_1.train_data.reshape(-1,784).float()/255.)
target = torch.Tensor(MNIST_1.targets)
data_txt = F.one_hot(target, 10)
data_txt = data_txt.type(torch.FloatTensor)

#Define parameters
input_dims=[784, 10]
max_epochs = 50
batch_size = 2000
latent_dim = 20

#Define model
mmvae = mmVAE(
        cfg="./config/MNIST_image_text.yaml",
        input_dim=input_dims,
        z_dim=latent_dim)

#Train the model
print('fit mmvae')
mmvae.fit(data_img, data_txt, max_epochs=max_epochs, batch_size=batch_size)

#Predict reconstructions
pred = mmvae.predict_reconstruction(data_img, data_txt)

#indices: view 1 latent, view 1 decoder, sample 21
pred_sample = pred[0][0][20]
data_sample = data_img[20]

fig, axarr = plt.subplots(1, 2)
axarr[0].imshow(data_sample.reshape(28,28))
axarr[1].imshow(pred_sample.reshape(28,28))
\end{minted}

\subsection*{Documentation}

In addition to the description of the \texttt{multi-view-AE} package provided here, documentation is available\footnote{https://multi-view-ae.readthedocs.io/en/latest/}  which serves as both guides to the \texttt{multi-view-AE} package and educational material for multi-view autoencoder models.

\begin{figure}[htb]
     \centering
         \includegraphics[width=\textwidth, scale=0.6]{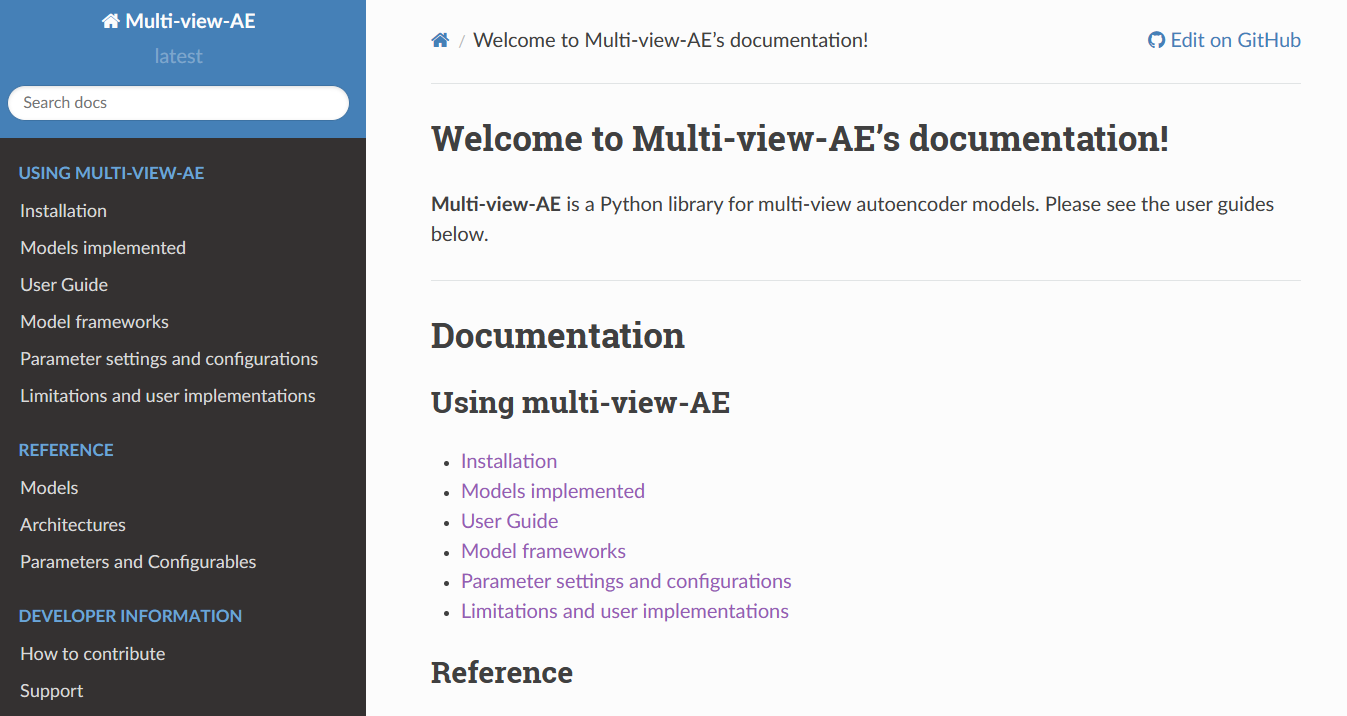}
        \caption{Screenshot of the homepage of the readthedocs documentation for \texttt{multi-view-AE}.}\label{fig:Documentation}
\end{figure}
\begin{figure}[htb]
     \centering
         \includegraphics[width=\textwidth, scale=0.6]{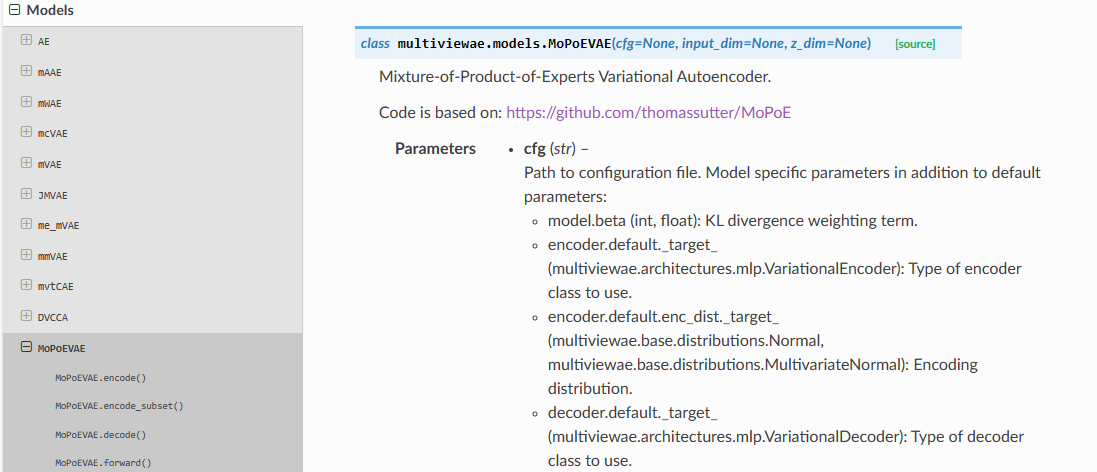}
        \caption{Screenshot of the readthedocs documentation with the model specific parameters for a model implemented in the \texttt{multi-view-AE} package.}\label{fig:Documentation_specific}
\end{figure}

\subsection*{Testing and parameter validation}

\texttt{multi-view-AE} contains extensive testing to ensure code quality and correct functionality of the models with a code coverage of 83\% at time of writing. \texttt{multi-view-AE} also contains validation functions to ensure the user inputs parameters of the correct type and combination. If an incorrect parameter combination is given, an error or warning is raised along with an explanation of the problem. An excerpt of validation code showing valid parameters and flags for invalid parameter combinations is shown below.

\begin{minted}
[
frame=lines,
framesep=2mm,
baselinestretch=1.2,
bgcolor=LightGray,
fontsize=\footnotesize,
linenos
]
{python}
from schema import Schema, And, Or, Optional, SchemaError, Regex

SUPPORTED_JOIN = [
            "PoE",
            "Mean"
        ]
   
config_schema = Schema({
    "model": {
        "save_model": bool,
        "seed_everything": bool,
        "seed": And(int, lambda x: 0 <= x <= 4294967295),
        "z_dim": int,
        "learning_rate": And(float, lambda x: 0 < x < 1),
        "sparse": bool,
        "threshold": Or(And(float, lambda x: 0 < x < 1), 0),
        Optional("eps"): And(float, lambda x: 0 < x <= 1e-10),
        Optional("beta"): And(Or(int, float), lambda x: x > 0),
        Optional("K"): And(int, lambda x: x >= 1),
        Optional("alpha"): And(Or(int, float), lambda x: x > 0),
        Optional("private"): bool,
        Optional("join_type"): eval(return_or(params=SUPPORTED_JOIN,
        msg="model.join_type: unsupported or invalid join type"))
    },

\end{minted}

\subsection*{Parameter settings}

The \texttt{multi-view-AE} package uses the Hydra API \citep{Yadan2019} for configuration management. \texttt{Hydra} is an open-source Python framework designed to simplify the development of Python projects. Its key feature is the ability to create a hierarchical configuration of Python objects which is easy to override, either through configuration files or the command line. In the \texttt{multi-view-AE} package, most parameters are set in a configuration file and are loaded into the model object by \texttt{Hydra}. The combination of \texttt{Hydra} with the modular structure of models in the package makes it easy for the user to replace model elements with other available implementations by editing the relevant section of the configuration file. Users are also able to specify their own functions and classes in the configuration file. These will often need to follow a particular structure as specified in the documentation. An example of a user implemented configuration file is provided below.

\begin{minted}
[
frame=lines,
framesep=2mm,
baselinestretch=1.2,
bgcolor=LightGray,
fontsize=\footnotesize,
linenos
]
{yaml}
# @package _global_

model:
  z_dim: 64
  s_dim: 30
  learning_rate: 0.00001
  
encoder:
  default:
    _target_: multiviewae.architectures.mlp.VariationalEncoder
    hidden_layer_dim: [256, 256]
    bias: True
    non_linear: True
    enc_dist:
      _target_: multiviewae.base.distributions.Normal

decoder:
  default:
    _target_: multiviewae.architectures.mlp.Decoder
    non_linear: True
    hidden_layer_dim: [256, 256]
    bias: True
    dec_dist:
      _target_: multiviewae.base.distributions.Bernoulli

\end{minted}

\section{Discussion}
In this work, we started by discussing the basis of multi-view autoencoders and different characterisations and frameworks. We went on to introduce a number of multi-view autoencoders implemented in the \texttt{multi-view-AE} package. Each model within the \texttt{multi-view-AE} package was described based on its defining characteristics, using consistent notation across models. We described the structure of the \texttt{multi-view-AE} package, highlighting key features and functionality. Our benchmarking results indicated that our implementations demonstrate comparable or improved performance compared to previous work. The design of \texttt{multi-view-AE} prioritises flexibility and user-friendliness, with a modular structure and API design that ensures it is accessible and intuitive for both beginners and experienced researchers.

\bibliographystyle{plainnat}  
\bibliography{references}  

\end{document}